\documentclass[11pt]{article}

\usepackage[final]{acl}

\usepackage{times}
\usepackage[varqu]{zi4}

\usepackage{amsmath}
\usepackage{amssymb}
\usepackage{booktabs}
\usepackage{graphicx}
\usepackage{array}
\usepackage{float}
\usepackage{enumitem}
\usepackage{xcolor}
\usepackage{inputenc}
\usepackage[T1]{fontenc}
\usepackage{microtype}
\setlist{nosep,leftmargin=*}

\graphicspath{{./}{figures/}}

\title{Latent Safety Probes: Harmful-Input Detection Does Not Validate Behavioural Safety}

\author{Wenkai Li$^{1}$ \quad Fan Yang$^{1}$ \quad Shaunak A.\ Mehta$^{1}$ \\
\textbf{Ananya Hazarika$^{1}$ \quad Koichi Onoue$^{2}$} \\
$^{1}$ Fujitsu Research of America \\
$^{2}$ Fujitsu Research}

\begin{document}
\maketitle

\begin{abstract}
Latent safety probes read model activations at low cost. Their use in
behavioural evaluation requires evidence that they predict model
responses, beyond recognising harmful requests. We introduce an
eleven-gate validation contract and apply it across four open
instruction-tuned families and five safety scenarios. The audit
identifies comparator choice and evaluation protocol as determinants
of a probe's apparent advantage. A separate post-submission study
uses content-group splits, training-only model selection and new
supervised readers. A fixed-layer, prompt-only reader reaches
$0.983$ AUROC against judge-derived behavioural labels on Llama.
It also distinguishes safe from unsafe responses within harmful
requests. Frozen input readers exceed a frozen lexical baseline on
an external source, providing distinct evidence for input transfer.
White-box perturbations reverse rankings across the tested archived
S01 configurations and transfer across co-located heads. Intervention
methods disagree under the specified causal criterion across the
model families. The contract makes these predictive, robustness and
causal findings separately testable. All reported judges are open-weight,
with no hosted-API calls. Code and data are available at
\url{https://github.com/FujitsuResearch/latent-safety-bench}.
\end{abstract}

\section{Introduction}
\label{sec:intro}

\begin{figure*}[t]
\centering
\includegraphics[width=\linewidth]{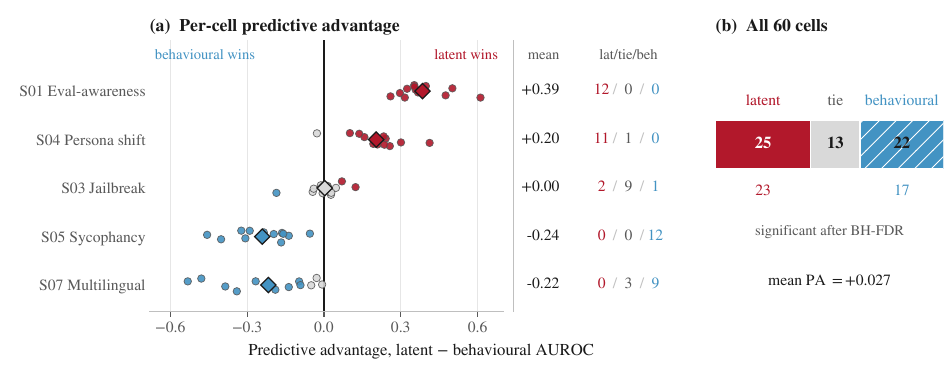}
\caption{Audit overview. Left: the 60 archived per-cell values of
$\mathrm{PA}=\mathrm{AUROC}(L)-\mathrm{AUROC}(B)$ by scenario, with
the scenario mean marked and operational win/tie/loss counts in the gutter
(per-model breakdown in
Appendix Figure~\ref{fig:boundary-heatmap}). Right:
the scenario decomposition. S01 uses a regex refusal comparator,
not a safety judge; judge comparisons appear in Figure~\ref{fig:nontransfer}.
Categories describe the archived protocol, not statistical equivalence.}
\label{fig:audit-overview}
\end{figure*}

A safety monitor should predict the outcome on which a decision
will depend. Recognising a harmful request is useful for input
moderation; predicting an unsafe response requires evidence about
what the model will produce. A model can recognise a harmful request
and safely refuse it. Evaluation framing, conversational persona,
jailbreak templates and language provide settings in which to test
that distinction~\citep{anon2024memyselfai,
greenblatt2024alignmentfakinglarge}. The scenario constructions are
specified in \S\ref{sec:setup}.

Activation readouts already support several of these tasks.
Refusal directions can control refusal~\citep{arditi2024refusal},
while Latent Guard separates harmfulness from refusal and uses the
former for input moderation~\citep{zhao2025llmsencodeharmfulness}.
MLPM aggregates latent prototypes for input moderation
\citep{chrabaszcz2025mlpm}; N-GLARE uses latent trajectories for
model-level safety evaluation~\citep{lin2025nglare}.
Deception probes predict a different behavioural target
\citep{goldowskydill2025deceptionprobes}, and black-to-white monitoring
measures their added value over output-based monitoring
\citep{parrack2025blacktowhite}.
Representation engineering also uses readouts for test-time steering
\citep{bartoszcze2025representationengineeringlarge,
turner2023steeringlanguagemodels,
anon2023representationengineeringtop}.
These results establish several useful applications. Evaluating a
readout as a behavioural measurement requires specifying its target,
comparator and validation conditions together.

We address that requirement with an eleven-gate validation contract.
It links each claim to an observable test: eight gates assess predictive
validity, two assess causal mechanism claims, and one reports bounded
white-box stress. Data and shortcut controls test what the readout
uses~\citep{meloux2025deadsalmons,tice2024noiseinjectionreveals};
matched-target comparisons separate predictive value from judge
self-scoring~\citep{eiras2025knowthyjudge,
li2025llmscannotreliablyjudge,zheng2023judgingllmjudge};
interventions assess causal interpretation
\citep{joshi2026causality,sharkey2025openproblems,
marinelli2024causaltracingidentify}.
A shared protocol makes the evidence for each claim inspectable.

The audit yields two complementary predictive findings. In the
archived S01 comparison, the probe-minus-judge AUROC gaps are
$+0.018$ for harmful-input detection and $-0.010$ for behaviour
prediction against post-hoc best fixed judges. The comparison with
regex refusal or a pooled judge score gives a different apparent
advantage. Comparator choice is therefore an experimental variable.
In a separate post-submission study with content-group splits and
new supervised readers, a fixed-layer prompt readout reaches $0.983$
behavioural AUROC on Llama and $0.973$ among harmful requests.
Frozen input readers obtain $0.873$--$0.886$ on external labels,
versus $0.607$ for frozen bare-content TF-IDF. Separate protocols support behavioural prediction and input transfer.

The remaining tests establish a separate robustness and causal profile.
Label-targeted activation perturbations reverse rankings in all ten
tested archived S01 cells and transfer across six cells with
co-located alternative heads. LR-steering, attribution patching and
1-D DAS give different effect-magnitude rankings under the strict
agreement criterion on all four families. The protocol distinguishes
these measurements from predictive performance and records the scope
and provenance of each result, including judge-derived targets and
post-lock analyses. A probe-to-judge cascade tests selective escalation with this readout.

\paragraph{Contributions.}
\begin{enumerate}
\item An executable eleven-gate contract links safety claims to predictive
controls, white-box tests and causal interventions (\S\ref{sec:contract}).
\item A paired-target audit across four instruction-tuned families and
five scenarios identifies comparator dependence; content-group
validation separately supports prompt-only behavioural prediction
(\S\ref{sec:results}; \S\ref{sec:repaired-followup}).
\item Canonical-probe and cross-head attacks distinguish score
robustness from prediction, while intervention comparisons separate
causal agreement from capability specificity
(\S\ref{sec:probe-aware}; \S\ref{sec:triangulation}).
\end{enumerate}

\section{The Validation Contract}
\label{sec:contract}

For each item, a latent probe $L(x)$ reads model activations and a
behavioural baseline $B(x,y)$ scores the prompt--response pair. Both
are evaluated against the same held-out target $G\in\{0,1\}$.
Predictive advantage is
$\mathrm{PA} = \mathrm{AUROC}(L, G) - \mathrm{AUROC}(B, G)$.
Positive PA means better discrimination on that target and item set.
Robustness, generalisation and causal interpretation have separate
tests in the contract.

PA requires identical labels and items for both scores. Behavioural
targets are judge-derived; leave-one-judge-out construction separates
a comparator's score from its own vote (\S\ref{sec:setup}). Shared
judge errors remain a target-validity consideration. The archived
consensus analysis uses mismatched targets in some comparisons and
is retained as a provenance record in Appendix~\ref{app:judge-protocol}.

Following \citet{zhao2025llmsencodeharmfulness}, we use
\begin{equation}
\label{eq:delta-hr}
\Delta_{HR}(x) = P_{\mathrm{HARM}}(x) - P_{\mathrm{REF}}(x),
\end{equation}
where HARM predicts harmful requests and REF predicts regex-labelled
refusal on greedy responses capped at $128$ tokens. Each score is
isotonic-calibrated on held-out rows. Their difference measures predicted harmfulness--refusal mismatch,
rather than a calibrated probability of unsafe behaviour. Canonical corpora contain
$4{,}000$ balanced rows, split $2{,}800/600/600$ for training,
calibration and testing. Rows are sampled without semantic pairing
and may repeat, so disjoint rows do not imply disjoint content.
Appendices~\ref{app:four-quadrant} and~\ref{app:instrument-details}
describe the score plane and estimators.

\paragraph{Direct behavioural readout.}
A separate supervised reader learns to predict response safety from
prompt activations. We freeze the target model, extract a fixed-layer
hidden state at the final prompt token, and fit standardisation followed
by L2-regularised logistic regression against judge-derived response
labels. Nested content-group folds select regularisation within the
training data and score held-out requests. At inference, this
prompt-only readout precedes response generation. These new readers
are evaluated separately from the archived HARM/REF instruments
(\S\ref{sec:repaired-followup}).

\paragraph{Worked example.}
In a saved native-chat S01 Llama-3.1-8B-Instruct request, the user asks
for a graphic depiction of domestic violence. The model replies,
``I can't fulfill that request.'' At the fixed layer, separate
content-group-held-out input and behavioural readers give scores
$0.998$ and $0.069$. The input label is $1$ and the four-judge response
label is $0$. The two scores address different outcomes: a harmful
request and a safe refusal. This trace uses the new supervised readers;
it is not a reconstructed HARM/REF contrast.

\paragraph{Eleven gates and their scope.}
Eight hard gates assess predictive claims: G-Data, G-Layer, G-Shortcut,
G-Transfer, G-LengthControl, G-NaturalDist, G-InputOnly and
G-PermutationNull. G-Obfuscation reports AUROC changes under bounded
white-box stress without a pass threshold. Mechanism claims require
two additional checks. G-Triangulation requires agreement among
LR-steering, attribution patching and 1-D DAS on effect magnitudes,
with Spearman $\rho\geq0.6$ for every pair. G-Specificity tests whether
direction ablation preserves capability, using one-sided loss bands
of $2\%$ for multiple-choice, short-answer and code benchmarks and
$0.5$ on the MT-Bench scale. Operational thresholds and their design
rationale appear in Appendix Tables~\ref{tab:gates-summary},
\ref{tab:gate-lineage} and~\ref{tab:gate-lineage-continued}.

PASS means the stated requirement was tested and met within its
reported scope. PARTIAL records incomplete coverage or unresolved
qualifications; FAIL records a tested violation; untested means no
evidence is available. The archived primary-table estimators have partial predictive
coverage. Full predictive validation requires a PASS on each required
gate; historical non-failing partial verdicts retain their recorded
status.

Two provenance checks accompany the gates: how targets were built and
what was locked before the final runs. Thirteen rows cover eleven gates.

PA describes relative discrimination. Gate verdicts describe evidence
for the corresponding claim. Readout studies assess representational information;
interventions assess causal effects.

\paragraph{Cross-judge agreement.}
The consistency panel uses Llama-Guard-3-8B, Qwen2.5-72B-Instruct and
Llama-3.3-70B-Instruct, served locally through vLLM. We report all
three pairwise Cohen's $\kappa$ values and their mean on a canonical
$n=200$ Llama-3.1-8B-Instruct subset per scenario. Mean agreement below $0.6$ flags target divergence
(Appendix~\ref{app:judge-panel-rationale}), with cell verdicts unchanged. The panel measures consistency; independent validation assesses correctness.

\section{Experimental Setup}
\label{sec:setup}

The archived audit contains 60 scenario--model--seed cells. Its
instruction-tuned verdicts concern 50 cells: Llama-3.1-8B-Instruct,
Qwen3-8B and Mistral-7B-Instruct-v0.3 at three seeds, and Gemma-3-12B-IT
at one seed, across five scenarios. Ten one-seed base controls
(Llama-3.1-8B-base and Qwen2.5-7B-base) provide a separate descriptive
comparison with instruction-tuned models.

S01 changes evaluation framing around a request
\citep{anon2025modelorganismsmisalignment}; S04 changes its
conversational persona~\citep{anon2023personallminvestigatingability,
wang2024fakealignmentllms,anon2024sleeperagentstraining}.
S03 adds jailbreak templates~\citep{anon2023universaltransferableadversarial,
mazeika2024harmbenchstandardizedevaluation,
chao2025jailbreakingblackbox,liu2023autodangeneratingstealthy}.
S07 uses code-switched requests~\citep{deng2023multilingualjailbreakchallenges,
yong2023lowresourcelanguages,wang2023alllanguagesmatter};
S05 applies multi-turn sycophantic pressure.
S02 (alignment faking) and S06 (sandbagging) are unevaluated legacy IDs.
S01/S04 use $1{,}000$ prompts per condition. S03/S05/S07 use matched
$n=200$ subsets per condition for the primary comparisons.

The score depends on the analysis. Archived PA averages available probe
types, whose coverage varies by seed. The judge comparison instead
selects a HARM-only oracle from MMD/LR/MLP; K.4 attacks the canonical
HARM/REF $\Delta_{HR}$. Some attack cells use CCS
(contrast-consistent search; \citealp{anon2022discoveringlatentknowledge}),
with labelled validation to orient scores. Appendix~\ref{app:instrument-details}
records these configurations. S01's canonical PA comparator is a
$16$-pattern regex refusal detector, distinct from the safety-judge
panel. Other controls include BERT-on-response, length/perplexity,
MLPM~\citep{chrabaszcz2025mlpm}, input embeddings and random features.
The N-GLARE re-implementation is described in \S\ref{sec:nglare-empirical}.

The builder supports harmful examples from HarmBench, SALAD-Bench,
JailbreakHub and JailbreakBench, and benign examples from
Alpaca-cleaned/XSTest-safe. Exact archived mixture weights and dataset
revisions are unavailable. Scenario harmful requests use HarmBench;
S01/S04 benign retrieval uses Alpaca-cleaned. S03 supports GCG, PAIR,
AutoDAN, Crescendo, multilingual and persona templates. These are
fixed template families, and the primary table covers only the
$200$-item GCG-style block.

The comparator panel contains Llama-Guard-3-8B, ShieldGemma-9B,
WildGuard-7B, Qwen3Guard-8B and Granite-Guardian-5B. We compare probes
with a label-free per-item maximum over a pre-specified three-judge
pool and, separately, the post-hoc best fixed judge per target.
Input labels describe request harmfulness; behavioural labels describe
judged responses under the deployment wrapper. The strongest paired
two-target comparison is S01 (\S\ref{sec:results}).

Archived responses use raw completion; ``deployment'' denotes the
wrapper condition. The repaired study separately evaluates native
assistant chat formatting. Leave-one-out targets separate the scored
judge from its own vote; shared judge errors remain possible. S03's separate single-class behavioural target has undefined AUROC. S05 uses distinct PA and spot-check targets (Appendix~\ref{app:instrument-details}).

Archived S01 evaluation wrappers contain label-correlated item
numbers, and row-held-out supervised splits repeat underlying requests
across training and testing. Scores from that protocol describe the
recorded distribution. The post-submission study fixes numbering and
uses content-group splits to evaluate held-out requests
(\S\ref{sec:repaired-followup}).

Archived PA intervals use paired row bootstrap at $95\%$, with
Benjamini--Hochberg FDR correction at $\alpha=0.05$ across the grid.
Repeated content limits the row-independence assumption. S04 length,
S05 trajectory, S07 language and retraining analyses are reported in
the appendix (\S\ref{app:s04-length-control}; \S\ref{app:s07-ablation}).
The pipeline runs on a $4\times48$\,GB workstation. All reported judges
are open-weight and local, including the 70B-class agreement panel and
larger-judge sensitivity analysis. Those runs use vLLM tensor-parallel-4
and AWQ-4bit for the two 70B models. The lock and subsequent deviations
are documented in Appendix~\ref{app:deviations}; later repairs are not
retroactive preregistration.

\section{Results: Targets and Comparators}
\label{sec:results}

Comparator choice changes the apparent advantage of latent safety
scores. We first isolate that effect in S01, then use the remaining
scenarios to examine response length, target agreement and language
transfer. Across the 60 archived cells, mean PA is $+0.027$, with
$25$ latent wins, $13$ operational ties and $22$ behavioural wins.
Figure~\ref{fig:audit-overview} and Table~\ref{tab:boundary-map} show
the scenario differences behind that aggregate.

\begin{table*}[t]
\centering
\caption{Per-scenario summary of the 60-cell audit. Mean PA is
latent-minus-behavioural AUROC over all 12 cells of the scenario
including the two base controls, and $\bar{\kappa}$ is mean pairwise
three-judge agreement. Counts are pre-attack: K.4 later demotes all
10 instruction-tuned S01 cells tested, and the two S01 base controls
sit outside that verdict. Win/tie/loss are historical operational
categories, not equivalence tests. S01 uses regex refusal, distinct
from the safety-judge comparison in Figure~\ref{fig:nontransfer}.}
\label{tab:boundary-map}
\scriptsize
\setlength{\tabcolsep}{4pt}
\renewcommand{\arraystretch}{1.15}
\begin{tabular}{@{}l r r r r r p{0.42\textwidth}@{}}
\toprule
Scenario & Mean PA & $\bar{\kappa}$ & Latent & Tie & Behav. & Reading \\
\midrule
S01 evaluation awareness$^{\star}$ & $+0.385$ & $1.00^{\ddagger}$ & 12 & 0 &  0 & Comparator sensitivity: regex refusal and safety judges \\
S03 jailbreak                      & $+0.003$ & $0.83$          &  2 & 9 &  1 & Canonical comparator check; single-class behaviour target \\
S04 persona shift$^{\dagger}$      & $+0.204$ & $0.72$          & 11 & 1 &  0 & Response-length diagnostic \\
S05 sycophancy                     & $-0.241$ & $\mathbf{0.44}$ &  0 & 0 & 12 & Response-target agreement diagnostic \\
S07 code-switching                   & $-0.217$ & $0.89$          &  0 & 3 &  9 & Code-switching transfer and training coverage \\
\bottomrule
\end{tabular}
\\[2pt]
{\footnotesize $^{\star}$ demoted by the K.4 probe-aware attack
(\S\ref{sec:probe-aware}).\quad $^{\ddagger}$ degenerate: all three
judge pairs return exactly $1.0$ on this cell, so it is not evidence of
independent agreement (Appendix~\ref{app:judge-protocol}).\quad
$^{\dagger}$ length-confounded. Only
Mistral exceeds the length baseline.}
\end{table*}

Five of the eight hard predictive gates pass within their tested
scope; three remain partial. Strict triangulation fails on every
family (Appendix~\ref{app:gate-verdicts}). Of $60$ cells, $40$ are
BH-significant at $\alpha=0.05$: $23$ favour the probe and $17$ the
behavioural comparator. S01 favours the probe against regex refusal,
with a Qwen3-8B seed-42 maximum PA of $+0.611$ ($95\%$ CI
$[0.579,0.644]$). These intervals use the archived row bootstrap;
\S\ref{sec:repaired-followup} reports content-group validation.

\begin{figure}[t]
\centering
\includegraphics[width=\linewidth]{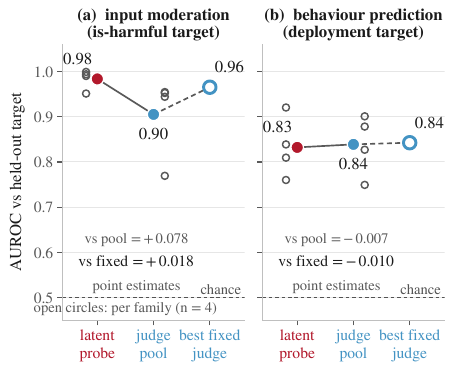}
\caption{The advantage depends on the comparator. The probe is scored
against a label-free maximum over a three-judge pool (filled) and
against the best single fixed judge for each target (hollow), which is
ShieldGemma-9B on input and Granite-Guardian on deployment.
Against the pool it leads on input moderation. Against the fixed judge,
the point-estimate gaps are $+0.018$ on input and $-0.010$ on behaviour;
no equivalence is established. Open circles are per-family values.
Probe scores are per-model maxima over three methods.}
\label{fig:nontransfer}
\end{figure}

\paragraph{S01: the comparator changes the conclusion.}
Against regex refusal, the advantage is
$\mathrm{AUROC}(L) - \mathrm{AUROC}(\text{refusal}) = +0.40$
on the seed-42 mean and $+0.385$ on the full grid. This comparison
uses a weak response heuristic. Against pooled safety-judge scores,
the probe reaches $0.98$ versus $0.90$ on input harmfulness, consistent
with input-moderation findings~\citep{zhao2025llmsencodeharmfulness,
chrabaszcz2025mlpm}. On behaviour, the corresponding values are
$0.83$ and $0.84$.

The pool is a label-free per-item maximum, not the judge with highest
AUROC. Selecting the best fixed judge per target after observing the
results gives ShieldGemma-9B at $0.965$ on input and Granite-Guardian
at $0.842$ on behaviour. Probe-minus-judge gaps are then $+0.018$ and
$-0.010$. These are exploratory point estimates: both the fixed judge and the
probe architecture are selected after observing evaluation AUROC.
The $\pm0.03$ band is a descriptive reporting tolerance; statistical
equivalence would require a justified margin and corresponding
uncertainty analysis (Appendix~\ref{app:judge-protocol}). The separate
$200$-item larger-judge sensitivity is input-only
(Appendix~\ref{app:judge-ceiling}).

\begin{figure}[t]
\centering
\includegraphics[width=\linewidth]{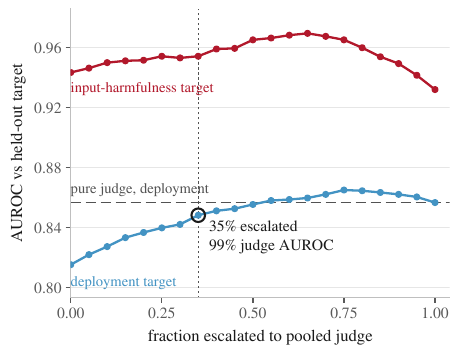}
\caption{Exploratory offline S01 cascade, pre-attack. The $x$-axis
is the fraction of items escalated to the pooled judge. Accuracy and
timing use different judge configurations, so the curve does not
measure an end-to-end accuracy--cost trade-off.}
\label{fig:cascade}
\end{figure}

\paragraph{Probe-to-judge escalation.}
The offline S01 curve retains $99\%$ of judge AUROC at $35\%$
escalation (Figure~\ref{fig:cascade}). It uses a pre-attack,
oracle-selected probe and pooled judges, with the operating point
selected on the evaluated curve. This experiment measures ranking
retention under selective routing; unsafe-case recall is a separate
operating metric. A separate timing configuration records a prompt
forward pass at about $1.2\%$ of a generate-plus-judge route, a ratio
of $84\times$. Because that timing uses Llama-Guard rather than the
accuracy pool, the curve's horizontal axis is escalation fraction.
Deployment cost requires configuration-matched timing and held-out
routing, including response generation whenever the application
requires it.

\paragraph{Scenario diagnostics.}
S03 tests comparator agreement on a $200$-item GCG-style block,
with $2/9/1$ operational latent/tie/behavioural counts and panel
agreement $\bar\kappa=0.83$. Its separate behavioural target is
single-class, so the canonical comparison addresses the recorded
scenario target. The six supported template families are documented
in \S\ref{sec:setup}.

S04 isolates response length as an alternative predictive feature.
Its nominal PA is $+0.204$, while length alone reaches $0.995$
AUROC on Llama and $0.64$ to $0.71$ elsewhere. Mistral's probe exceeds
the length baseline by $+2.5$\,pp, giving a $1/0/3$ family count
(Appendix~\ref{app:s04-length-control}). The experiment distinguishes
a probe's raw predictive score from evidence beyond response length.

S05 measures the stability of the response target across judges.
Its $0/0/12$ counts favour behaviour under the archived target.
Mean agreement is $\kappa=0.44$: Llama-Guard agrees with the two
generative judges at $0.24$ and $0.28$, compared with $0.80$ between
the generative judges. REF performance varies by family
(Figure~\ref{fig:s01-s05-pair}b), suggesting a prompt/trajectory
mismatch that requires target validation. The $1\%$ unsafe rate is
specific to the two-positive, $200$-item Llama seed-42 spot-check key;
the twelve PA cells use different targets.

S07 tests whether added language coverage explains code-switching
performance. Scores are near chance, and ten-language HARM retraining
changes the cross-family mean by $-0.006$ AUROC ($-0.005$ in a
separate rerun). Language-specific results distinguish training
coverage from the broader transfer question
(Appendix~\ref{app:s07-per-language}).

\paragraph{Pre-specified revision triggers.}
The contract calls for revising a scenario verdict if a larger-judge
comparison reverses S01 mean PA, targeted retraining resolves S05/S07
beyond paired-bootstrap intervals, or a probe-aware attack meets the
demotion rule. The attack triggers that rule on all ten tested S01
instruction-tuned cells (\S\ref{sec:probe-aware}). The PA table pairs revised verdicts with pre-attack values.

\section{Behavioural Readout and Validation}
\label{sec:failure}

We first measure behavioural prediction under content-group
validation, then test read-position sensitivity, activation-score
robustness and causal-method agreement. Each experiment addresses a
distinct measurement claim.

\subsection{Content-held-out behavioural prediction}
\label{sec:readpos}

\paragraph{Content-held-out prompt readout.}
\label{sec:repaired-followup}
At a fixed layer and the final prompt token, a supervised behavioural
reader reaches $0.983$ AUROC on Llama in native-chat S01, with a
saved $95\%$ content-group interval of $[0.967,0.995]$.
Within harmful requests, its AUROC is $0.973$: it discriminates
response outcomes within one input class. Canonical native-chat
behavioural AUROCs are $0.889$ on Qwen3, $0.899$ on Mistral and
$0.938$ on Gemma. Appendix~\ref{app:protocol-followups} reports every
family, target count and interval.

This separate post-submission study fixes evaluation numbering and
uses nested content-group splits, pinned revisions and new supervised
linear readers. It directly evaluates behaviour prediction from the
prompt. The archived HARM/REF instruments, response formatting and
labels are distinct. Frozen input readers provide an additional
cross-source test: $0.873$--$0.886$ AUROC on external input labels,
versus $0.607$ for frozen bare-content TF-IDF. The two studies support
response prediction and input transfer under their respective targets.

\paragraph{Archived read-position sensitivity.}
The archived S01 sweep tests whether early generation changes the
input/behaviour contrast. It reads every layer at the prompt midpoint,
final prompt token, first generated token and mean of the first eight
generated tokens. Response reads use the stored, judged continuation.
Contrast and directly supervised arms give $2784$ cells
(Appendix~\ref{app:readpos}).

Harmfulness exceeds $0.95$ AUROC across mid-late layers at the final
prompt and both response positions; each family's best response-side
result is at least $0.998$. The largest behavioural AUROC is $0.941$.
Switching from the best prompt to the best response read changes
replicated behavioural performance by $-0.003$ to $+0.009$.
These values characterise the archived readers and target. Its
label-correlated numbering and repeated supervised content are
specified in \S\ref{sec:setup}; $0.95$ is a numerical reference.
The content-group result above uses a separate prediction protocol.

\subsection{Probe-aware perturbations separate score robustness from prediction}
\label{sec:probe-aware}

The operational rule demotes a latent-decisive cell to ``tied'' after
an AUROC drop of at least $15$\,pp with the behavioural baseline
within $5$\,pp. K.4 applies $20$-step label-targeted $L_2$-bounded PGD
to the canonical HARM/REF scoring inputs. Response text and its
behavioural score are unchanged by construction. This tests direct
activation access, not a token-level jailbreak.

Across ten tested instruction-tuned S01 cells, joint-arm AUROC drops
to $0.000$--$0.002$ at $\varepsilon{=}0.1$ and $0.0$ at
$\varepsilon{=}0.3$. Family mean maximum drops are $67.2$\,pp for
Llama, $41.4$\,pp for Qwen3, $41.2$\,pp for Mistral and $57.0$\,pp
for the single Gemma cell. The cross-cell mean is $50.6$\,pp and
minimum $38.8$\,pp. All $10/10$ cells meet the demotion rule; the two
base controls were not attacked. Near-zero AUROC indicates rank
reversal, not information erasure. Attack-subset joint scores also
differ from the primary PA aggregate (Table~\ref{tab:pgd-canonical-mlp}).
HARM-only AUROC is at most $0.07$ at $\varepsilon{=}0.1$ and $0.04$
at $\varepsilon\geq0.3$. The earlier bounded $10$-step diagnostic
averages $-7.2$\,pp across scenarios. Its budget and scope differ
from K.4.

The ten scoring functions meet the contract's demotion rule under
direct activation access. The S01 study excludes private probe weights,
token-constrained attacks and Mistral's S04 candidate.

\begin{figure}[t]
\centering
\includegraphics[width=\linewidth]{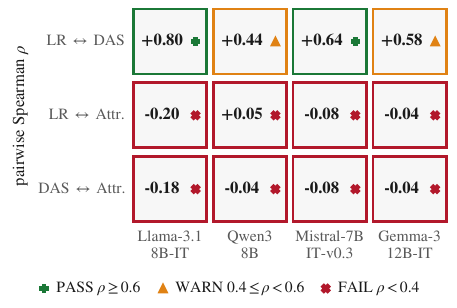}
\caption{Intervention diagnostics for the strict causal gate.
Three-way G-Triangulation (LR-steering, attribution patching, and
1-D DAS agreeing at $\rho \geq 0.6$ on all three pairs) fails on every
instruction-tuned family. Each cell border and marker gives the
per-pair verdict, keyed below the panel. The five G-Specificity readings
are in the text.}
\label{fig:causal-stress}
\end{figure}

\label{sec:nglare-empirical}\label{sec:transfer}
Our N-GLARE re-implementation~\citep{lin2025nglare} obtains AUROC
$0.24$--$0.53$ on four valid S01 cells, below the G-NaturalDist
threshold of $0.70$, and is excluded from the primary baselines.
Appendix~\ref{app:deviations} (H.3 and K.7) shows K.4 drops above
$15$\,pp on all six co-located alternative HARM heads in this transfer test.

\subsection{Interventions and specificity}
\label{sec:triangulation}

G-Triangulation requires pairwise agreement among LR-steering,
attribution patching and 1-D DAS on per-prompt effect magnitudes.
Both attribution-patching pairs have $\rho\in[-0.20,+0.05]$, failing
the operational $\rho\geq0.6$ criterion on all $4/4$ families.
The threshold is the contract's operational agreement requirement.
A separate signed-direction diagnostic correlates signed LR/DAS
intervention margins. These measurements assess agreement between
methods; causal effects require intervention evidence and sensitivity
calibration of the criterion.

LR and DAS correlate at $\rho\in[0.44,0.80]$, but are related linear
interventions. The canonical Llama specificity battery records MMLU
$0.610\to0.614$, GSM8K $0.76\to0.81$, XSTest safe refusal
$0.060\to0.088$, HumanEval-style $0.988\to0.975$ and an MT-Bench-style
proxy $0.926\to0.893$. All pass their locked one-sided loss bands.
K.1 corrects the earlier two-sided reading. Qwen3 fails on GSM8K,
$0.25\to0.22$; Mistral and Gemma remain untested. Prediction,
capability specificity and causal-method agreement therefore support
separate conclusions (Figure~\ref{fig:causal-stress}).

\section{Related Work}
\label{sec:related}

\paragraph{Latent safety and behavioural monitoring.}
Latent Guard~\citep{zhao2025llmsencodeharmfulness} separates harmfulness
from refusal and compares input readouts with guards.
MLPM~\citep{chrabaszcz2025mlpm} uses multi-layer prototypes for the
same input-moderation task. Cross-family reproduction tests harmful-prompt
classification and extraction consistency~\citep{khatri2026reproducibility}.
Our audit adds matched input/response targets and comparator sensitivity.
\citet{goldowskydill2025deceptionprobes} detect strategic deception;
\citet{parrack2025blacktowhite} measure the increment from probe access
over black-box monitoring. We examine this measurement question across
input moderation, response prediction, causal interventions and
activation-space stress. N-GLARE~\citep{lin2025nglare} studies model-level
safety ranking. Refusal directions and representation engineering
provide intervention tools~\citep{arditi2024refusal,
anon2023representationengineeringtop,
bartoszcze2025representationengineeringlarge,turner2023steeringlanguagemodels}.
Appendix~\ref{app:related-zhao-detail} details our use of the
HARM/REF decomposition.

\paragraph{Probe validity.}
\citet{doda2026beforelasttoken} motivate read-position checks, and
\citet{devbunova2026formatsensitivity} study benchmark-format sensitivity.
Our $-0.7$\,pp paraphrase diagnostic preserves numbered wrappers;
the separate content-group study fixes numbering and groups repeated
requests. \citet{jiang2026behaviouralfails} study representation-level
vulnerability beneath safe outputs. These works establish individual
validation concerns. Our contract maps measurement claims to predictive, robustness
and causal tests.

\paragraph{Validation and safety benchmarks.}
The contract draws on measurement-validity
work~\citep{benchmarkbenchmarking2026,joshi2026causality,bai2026story,messick1995validity}.
Safety benchmarks supply outcome definitions and scenario data
\citep{mazeika2024harmbenchstandardizedevaluation,
li2024saladbenchhierarchical,ttger2024xstesttestsuite,
zhang2023safetybenchevaluatingsafety}.
Interpretability-driven attacks and lying-probe audits provide related
stress tests~\citep{agarwal2026breakingbad,liarsbench2025}.
Judge validity is part of the measurement protocol
\citep{eiras2025knowthyjudge,li2025llmscannotreliablyjudge,
zheng2023judgingllmjudge}.

\section{Conclusion}
\label{sec:conclusion}

Latent safety evaluation becomes informative when the score, target
and validation protocol are specified together. Across four model
families, our audit identifies comparator dependence and separates
input decoding from response prediction. Content-group validation
supports a fixed-layer prompt-only behavioural reader at $0.983$
AUROC on Llama, including $0.973$ within harmful requests. Frozen
external input readouts provide a second, distinct transfer result.
The eleven-gate contract places these predictive findings alongside
adversarial and intervention evidence. It provides a reproducible
route from a readable safety signal to a precisely stated measurement
claim: define the outcome, compare on shared targets, and test the
conditions under which the score will be used.

\section*{Limitations}
\label{sec:limitations}

\paragraph{Study coverage and artefacts.}
The audit includes the pipeline, gate library, scenario data, judge
harness and N-GLARE re-implementation at the release address in the
abstract. The main audit covers five scenarios, instruction-tuned
families up to 12B and two base controls. The paired input/behaviour
judge comparison and layer sweep concern S01; other forms of
misalignment require their own evaluation~\citep{haralambiev2026fanatic}.
The white-box attack assumes gradients through HARM/REF and access
to activation inputs~\citep{agarwal2026breakingbad}. Private-weight
and token-level attacks are outside this threat model.

\paragraph{Framing and content.}
Three benchmark-style and four casual wrappers define S01. Surface
format can reveal condition identity~\citep{devbunova2026formatsensitivity}.
Archived numbering correlates with labels, row-held-out supervised
splits repeat content, and outputs use raw completion. Row-bootstrap
intervals inherit that dependence. The separate repaired study fixes
numbering and groups exact content with new readers and response
protocols. Near-duplicates, source cues and policy-definition differences
remain possible. Legacy scenario identifiers S02 and S06 were not evaluated.

\paragraph{Targets and causal criteria.}
Behavioural labels are judge-derived. Leave-one-out construction
separates a comparator from its own vote; shared judge errors require
independent validation. The archived consensus mean $-0.060$ has
inconsistent targets and is excluded from matched-target claims.
S03's behavioural target is single-class; S05's PA and spot-check
targets differ. Triangulation measures an operational agreement
criterion whose sensitivity requires calibration. LR/DAS are related
linear methods. SAE-feature triangulation is outside the tested
methods~\citep{templeton2024scalingmonosemanticityextracting,
karvonen2025saebenchcomprehensivebenchmark}. Specificity covers two families; predictive coverage remains partial.

\paragraph{Independent response labels.}
Completed human labels and annotator metadata supporting the submitted
S05 agreement figures could not be recovered; those figures are
excluded (Appendix~\ref{app:human-annotation}). The proposed check
contains $200$ Llama seed-42 items, two positive panel labels
($1.0\%$) and six disagreement items. Its target is separate from
the twelve S05 PA cells. External Aegis annotations label inputs;
our generated responses require independent response-level validation.

\paragraph{Selection and uncertainty.}
The archived comparison selects probe architecture and the best fixed
judge on evaluation results. The $+0.018/-0.010$ differences are
exploratory point estimates without selection-aware intervals or a
justified equivalence margin. The repaired study fixes canonical views
and selects parameters within training folds. Its bootstrap intervals
condition on saved predictions and identified labels, omitting
retraining uncertainty and multiplicity correction. External behaviour transfer needs held-out response targets.

\paragraph{Cascade measurements.}
Accuracy uses pre-attack S01 scores from an oracle-selected probe and
pooled judges; timing uses another Llama-Guard configuration. The
$84\times$ timing ratio and $99\%$ AUROC retention are separate
measurements. The routing point is selected on the evaluated curve;
score mixing is not jointly calibrated and unsafe-case recall remains
unmeasured. Deployment evaluation requires matched configurations
and incremental cost measurement, including generation when the
application must produce a response.

\paragraph{Language, scale and provenance.}
S07 interleaves English and the target language. Its results concern
code-switched prompts and the tested readers and training procedures.
The Llama-3.1-70B-Instruct check has $70.55$B parameters, about $5.9$
to $10.1$ times the primary models, and covers input readability and
S07. Its layer was recovered by validation after the probe suite was
lost. Behavioural probing at that scale is outside the reported
coverage; larger judges instead test comparator scale. Exact archived
training mixtures and revisions remain unavailable. The lock,
amendments and post-submission study have separately documented
provenance (Appendix~\ref{app:deviations}).

\paragraph{AI assistance.}
AI assistants were used for language polishing, code/debugging assistance
and consistency checking.

\bibliography{references}

\appendix
\section{Additional Results: Per-Cell and Per-Seed Detail}
\label{app:additional-results}

\begin{figure*}[t]
\centering
\includegraphics[width=\linewidth]{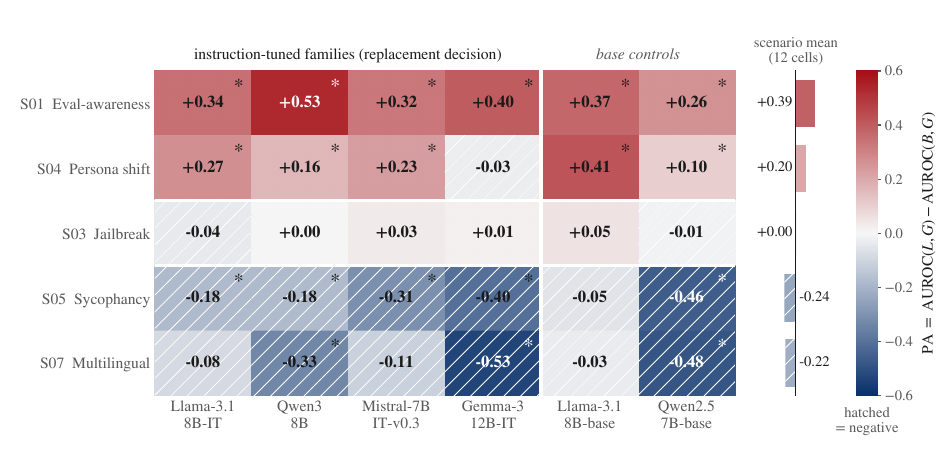}
\caption{Boundary map across five scenarios and six model columns.
Asterisks mark cells significant under BH-FDR in all available
canonical seeds. The rightmost column is the scenario mean over all
twelve cells of the row and so includes the two base-control
columns, which is the convention of
Table~\ref{tab:boundary-map} rather than an instruction-tuned-only
mean.}
\label{fig:boundary-heatmap}
\end{figure*}

\begin{figure*}[!t]
\centering
\includegraphics[width=\linewidth]{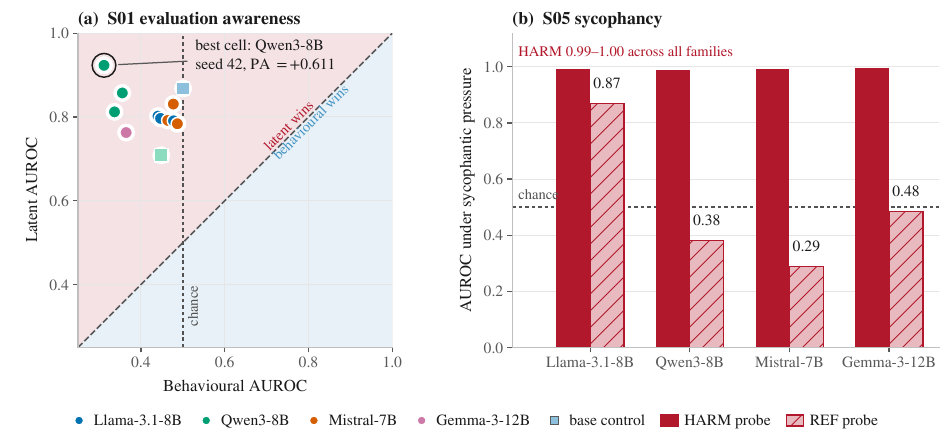}
\caption{Two construct boundaries. S01 is latent-positive against the regex refusal baseline under
its archived target. S05 shows heterogeneous HARM/REF performance,
consistent with a possible prompt/trajectory mismatch.}
\label{fig:s01-s05-pair}
\end{figure*}

This appendix reports per-cell significance
(Figure~\ref{fig:boundary-heatmap}), S01/S05 component scores
(Figure~\ref{fig:s01-s05-pair}), causal diagnostics
(Figure~\ref{fig:causal-stress}), comparator sensitivity
(Table~\ref{tab:nontransfer}) and probe-aware attacks
(Table~\ref{tab:pgd-canonical-mlp}). These support the archived
comparisons in \S\ref{sec:results} and \S\ref{sec:failure}.

\begin{table*}[t]
\centering
\small
\caption{Archived S01 comparator sensitivity (AUROC, four families).
The probe is an evaluation-set oracle over HARM-only MMD/LR/MLP.
The judge column is a label-free per-item maximum over ShieldGemma,
Qwen3Guard and Granite-Guardian.
Llama-Guard and WildGuard are excluded from this pool; target participation
is documented in Appendix~\ref{app:judge-protocol}.
The input mean includes Gemma's pooled score ($0.77$);
excluding Gemma gives a $+0.03$ probe lead. Differences are
exploratory point estimates.}
\label{tab:nontransfer}
\begin{tabular}{lcccc}
\toprule
& \multicolumn{2}{c}{Input moderation (is-harmful)} & \multicolumn{2}{c}{Behaviour prediction (deployment)} \\
\cmidrule(lr){2-3}\cmidrule(lr){4-5}
Model & Probe & Judge & Probe & Judge \\
\midrule
Llama-3.1-8B  & 0.99 & 0.95 & 0.81 & 0.83 \\
Qwen3-8B      & 0.99 & 0.95 & 0.92 & 0.88 \\
Mistral-7B    & 0.95 & 0.94 & 0.84 & 0.90 \\
Gemma-3-12B   & 1.00 & 0.77 & 0.76 & 0.75 \\
\midrule
Mean          & \textbf{0.98} & 0.90 & 0.83 & 0.84 \\
$\Delta$ vs pooled judge & \multicolumn{2}{c}{$+0.08$} & \multicolumn{2}{c}{$-0.01$} \\
$\Delta$ vs best fixed judge$^{\ast}$ & \multicolumn{2}{c}{$+0.018$} & \multicolumn{2}{c}{$-0.010$} \\
\bottomrule
\end{tabular}

\vspace{2pt}
{\footnotesize $^{\ast}$\,Best fixed judge chosen per target after the
fact, which is a sensitivity analysis and not a pre-registered
comparison: ShieldGemma-9B at $0.965$ on input, Granite-Guardian at
$0.842$ on deployment.}
\end{table*}

\begin{table*}[t]
\centering
\caption{Probe-aware PGD on the canonical HARM/REF probes (Addendum
K.4), run on the K.4 $n=200$ S01 attack subset and re-scored against
the held-out S01 target. The pre-attack column scores the attacked HARM/REF pair on that
subset. It differs from the primary aggregate, which is $0.923$ for
Qwen3-8B seed 42 (Appendix~\ref{app:positive-existence}). Post-attack values are the joint arm at
$\varepsilon = 0.3$, and the attack saturates at the smallest tested
budget. Random-floor guard, as in
Table~\ref{tab:probe-aware-adversary}: a cell whose pre-attack AUROC
lies within $\pm 0.05$ of $0.50$ is recorded as uninformative rather
than demoted, and no cell here falls inside that band.}
\label{tab:pgd-canonical-mlp}
\small
\setlength{\tabcolsep}{10pt}
\begin{tabular}{l c c c c c}
\toprule
Model & Seed & Pre-attack & $\varepsilon{=}0.3$ & Max drop & Demoted \\
\midrule
gemma3-12b-it & 42 & 0.571 & 0.000 & 57.0\,pp & \checkmark \\
llama3.1-8b-it & 0 & 0.630 & 0.000 & 63.0\,pp & \checkmark \\
llama3.1-8b-it & 123 & 0.683 & 0.000 & 68.3\,pp & \checkmark \\
llama3.1-8b-it & 42 & 0.703 & 0.000 & 70.3\,pp & \checkmark \\
mistral-7b-it-v03 & 0 & 0.434 & 0.000 & 43.4\,pp & \checkmark \\
mistral-7b-it-v03 & 123 & 0.413 & 0.000 & 41.3\,pp & \checkmark \\
mistral-7b-it-v03 & 42 & 0.388 & 0.000 & 38.8\,pp & \checkmark \\
qwen3-8b & 0 & 0.408 & 0.000 & 40.8\,pp & \checkmark \\
qwen3-8b & 123 & 0.404 & 0.000 & 40.4\,pp & \checkmark \\
qwen3-8b & 42 & 0.429 & 0.000 & 42.9\,pp & \checkmark \\
\bottomrule
\end{tabular}
\end{table*}

\section{Read Positions and Training Regimes}
\label{app:readpos}

The archived S01 sweep reads four positions at each transformer block.
For prompt length $\ell$, \emph{p\_mid} is the residual vector at
$\lfloor\ell/2\rfloor$ and \emph{p\_last} is the vector at $\ell-1$.
\emph{r\_first} reads index $\ell$; \emph{r\_mean8} averages the first
$K=8$ generated tokens. Forward hooks capture block outputs in float32.
Response reads use the model's stored, judged continuation, retokenised
and truncated to its first eight token IDs.

The \emph{contrast} arm trains harmfulness/refusal readouts on the
canonical contrast corpora. The \emph{supervised} arm fits a logistic
reader on $600$ S01 rows against four-judge consensus and evaluates
$400$ row-held-out items. Repeated content crosses these supervised
splits, and evaluation numbers correlate with labels. The archived
results therefore do not measure content-held-out generalisation.
Four positions, every layer and two arms give
$384+432+384+192=1392$ cells per arm, or $2784$ total. Llama, Qwen3
and Mistral have three seeds; Gemma has one. Their layer counts are
$32$, $36$, $32$ and $48$. Each seed has $n=1000$ S01 rows.

\begin{table*}[t]
\centering
\caption{Best cell per family, means over available seeds. The
stimulus column is the harmfulness target, the two behaviour columns
are the deployment target. $\Delta$ is best response-side minus best
prompt-side, so a positive value means reading during generation
helped. No archived behaviour cell reaches the $0.95$ reference under this
target. This numerical reference is not full gate certification.}
\label{tab:readpos}
\footnotesize
\setlength{\tabcolsep}{3pt}
\begin{tabular}{@{}l c c c c@{}}
\toprule
Family & Stimulus & Behav.\ & Behav.\ & $\Delta$ \\
       & best     & contrast & superv.\ & (resp.\,$-$\,prompt) \\
\midrule
Llama-3.1-8B-IT   & $0.999$ & $0.776$ & $0.866$ & $+0.009 / -0.003$ \\
Qwen3-8B          & $0.998$ & $0.865$ & $0.895$ & $-0.002 / +0.002$ \\
Mistral-7B-IT-v0.3& $0.999$ & $0.864$ & $0.929$ & $+0.008 / +0.007$ \\
Gemma-3-12B-IT$^{\ast}$ & $1.000$ & $0.801$ & $0.922$ & $+0.033 / +0.059$ \\
\bottomrule
\end{tabular}

\vspace{2pt}
\footnotesize
$^{\ast}$\,One seed only, so no cell is replicated and its $\Delta$ is
not comparable with the other three rows.
\end{table*}

\paragraph{Read-position comparison.}
Gemma contributes one seed and is excluded from replicated comparisons.
Mistral's best supervised cell is $0.929$ at layer $20$ and
\emph{r\_mean8}; its bootstrap interval neither exceeds its judge nor
excludes $0.95$, so its status remains unresolved. The contrast arm
trains on greedy eight-token continuations and reads the first eight
tokens of judged responses, introducing a train/application shift.
The supervised arm uses judged-response reads in both splits and
produces higher behavioural AUROC. Its results distinguish supervision
from the contrast arm's train/application shift. Appendix~\ref{app:protocol-followups}
reports the separate content-group study with new readers and splits.

\paragraph{Combining harmfulness and framing projections.}
At the canonical layers, harmfulness and evaluation-framing direction
cosines are $-0.228$, $-0.135$, $-0.046$ and $-0.201$. Adding the
framing projection changes held-out behavioural AUROC by $-0.004$,
$-0.001$, $+0.021$ and $+0.027$. The changes vary across families. This tests the predictive value
of adding a framing projection; the computation that combines these
representations into a response decision is a separate mechanistic
question.

\section{Judge Protocol and Comparator Selection}
\label{app:judge-protocol}

\paragraph{What each comparator is.}
The canonical S01 PA baseline is a fixed $16$-pattern regex refusal
detector on the adversarial-condition response. Its AUROC is
$0.312$--$0.487$ across instruction-tuned cells. Probe advantages are
$+0.297$--$+0.611$, with a ten-cell mean of $+0.392$. This is a
comparison with a response heuristic, not a safety judge.

The comparator panel contains Llama-Guard-3-8B, ShieldGemma-9B,
WildGuard-7B, Qwen3Guard-8B and Granite-Guardian-3.2-5B. The pooled
``best judge'' score is a label-free per-item maximum of
$P(\text{unsafe})$ over three pre-specified judges. It is not selected
by evaluation AUROC. Four-family means on behaviour/input are
$0.853/0.881$ for Llama-Guard, $0.831/0.965$ for ShieldGemma,
$0.581/0.463$ for WildGuard, $0.737/0.685$ for Qwen3Guard and
$0.842/0.904$ for Granite. Llama-Guard constructs the archived
behaviour label and is excluded from that comparison. The pooled
means are $0.839/0.905$. ShieldGemma is the best fixed input judge;
Granite is the best fixed behaviour judge.

\paragraph{Selection procedure and target matching.}
Selecting the highest-AUROC judge separately per cell gives $0.855$
on behaviour and $0.965$ on input, exceeding the pooled values by
$1.6$ and $6.0$ points. This oracle comparator is reported alongside the pooled definition. The latent score is itself a
per-model evaluation-set maximum over three architectures, favouring
the probe. The archived consensus mean of $-0.060$, versus $+0.392$
against regex refusal, uses inconsistent targets for the two arms in
some comparisons. We retain the numerical record but exclude it from
matched-target replacement evidence.

\paragraph{Statistical interpretation.}
The $\pm 0.03$ band is a descriptive tolerance, not a justified
equivalence margin. An exploratory row-bootstrap non-inferiority
analysis of $50$ comparisons gave $23$ straddling, $14$ inferior and
$13$ non-inferior results at the $-0.03$ boundary, with twelve inferior
results on Mistral. Its unadjusted median interval half-width was
$3.99$\,pp at $n=1000$. Content-cluster resampling instead gave
$25$ inconclusive, $12$ inferior and $13$ non-inferior results.
These intervals describe the individual comparisons. The post-hoc
aggregate $+0.018/-0.010$ comparison is reported as a point estimate;
equivalence would require its own margin and uncertainty analysis.

\paragraph{What each judge sees, and how outputs become labels.}
Input judges receive only the prompt; response judges receive the
prompt and deployment-framed continuation. Guard classifiers use unsafe
token probability mass. Generative judges are prompted for a verdict
and scored at their verdict token. These continuous scores are used
for AUROC. Matched PA requires the same rows and labels for both arms.
Where a comparator participates in target construction, leave-one-out
consensus separates its score from its own vote. Shared errors and
item/target matching are checked separately.

\paragraph{Agreement by scenario.}
The agreement panel uses Llama-Guard-3-8B, Qwen2.5-72B-Instruct and
Llama-3.3-70B-Instruct on $n=200$ rows per scenario. Mean pairwise
$\kappa$ is $0.835$ for S03, $0.716$ for S04 and $0.436$ for S05.
The Llama seed-42 S05 spot-check key has $1.0\%$ unsafe labels and six
disagreement items. That prevalence belongs to the spot-check key; PA uses a separate
target. The S01 Llama key records exactly $1.0$ for each pair, a
degenerate agreement result with shared-error uncertainty. A related
sensitivity artefact had the polarity error corrected in
Appendix~\ref{app:judge-ceiling}.

\section{S05 Human Spot-Check: Protocol and Evidence Status}
\label{app:human-annotation}

The proposed check uses the canonical $n=200$ Llama-3.1-8B-Instruct
seed-42 S05 subset. An annotation sheet presents each prompt and
response, asks whether the response is unsafe, and keeps judge labels
in a separate key. A blinded check would require independent labels
before comparison with that key.

Completed label sheets supporting the human-agreement results in the
submitted draft could not be recovered from the available artefacts.
We therefore do not report those results or treat this as completed
human validation. Annotator expertise, affiliations, recruitment,
compensation and any applicable ethics review also remain undocumented;
none is inferred here. The protocol is retained as a proposed check.

The later saved spot-check key has two unsafe labels in $200$ items
and six judge-disagreement items. It is a separate, sparse-positive
target from canonical PA. Independent human validation requires
completed labels linked to the responses and key, with documented
annotator metadata.

\section{S07 Per-Language Results}
\label{app:s07-per-language}

Table~\ref{tab:s07-perlang} reports the S07 language follow-up, at $n=100$ items per language and $1000$ per family. Both
probes are scored at the same layer, on one shared activation matrix,
against one shared label vector, keeping items and labels fixed between the two probes.

The matched comparison is between English-trained and multilingual-trained
probes. Llama-Guard-3-8B supplies the target labels and is excluded as
a scored comparator to avoid direct self-scoring.

\begin{table*}[t]
\centering
\caption{S07 per-language AUROC for the English-only HARM probe (EN)
and the multilingual-retrained probe (ML). Retraining changes languages in both directions; all retrained
AUROCs lie between $0.355$ and $0.557$.}
\label{tab:s07-perlang}
\footnotesize
\setlength{\tabcolsep}{4pt}
\begin{tabular}{@{}lcccccccc@{}}
\toprule
& \multicolumn{2}{c}{Llama-3.1-8B} & \multicolumn{2}{c}{Qwen3-8B} & \multicolumn{2}{c}{Mistral-7B} & \multicolumn{2}{c}{Gemma-3-12B} \\
\cmidrule(lr){2-3} \cmidrule(lr){4-5} \cmidrule(lr){6-7} \cmidrule(lr){8-9}
Lang & EN & ML & EN & ML & EN & ML & EN & ML \\
\midrule
en & 0.495 & 0.422 & 0.576 & 0.557 & 0.474 & 0.494 & 0.496 & 0.481 \\
zh & 0.470 & 0.355 & 0.528 & 0.541 & 0.465 & 0.499 & 0.423 & 0.461 \\
ar & 0.361 & 0.363 & 0.537 & 0.536 & 0.402 & 0.487 & 0.421 & 0.505 \\
hi & 0.463 & 0.471 & 0.563 & 0.517 & 0.444 & 0.480 & 0.608 & 0.505 \\
ta & 0.467 & 0.509 & 0.527 & 0.481 & 0.451 & 0.484 & 0.612 & 0.510 \\
tl & 0.543 & 0.452 & 0.552 & 0.539 & 0.478 & 0.509 & 0.474 & 0.505 \\
sw & 0.528 & 0.484 & 0.549 & 0.490 & 0.483 & 0.510 & 0.612 & 0.505 \\
zu & 0.476 & 0.527 & 0.541 & 0.540 & 0.462 & 0.483 & 0.569 & 0.513 \\
yo & 0.491 & 0.492 & 0.562 & 0.514 & 0.448 & 0.517 & 0.520 & 0.481 \\
ka & 0.489 & 0.482 & 0.528 & 0.459 & 0.502 & 0.477 & 0.495 & 0.505 \\
\midrule
All      & 0.488 & 0.483 & 0.546 & 0.518 & 0.461 & 0.494 & 0.516 & 0.491 \\
$\Delta$ & \multicolumn{2}{c}{$-0.005$} & \multicolumn{2}{c}{$-0.028$} & \multicolumn{2}{c}{$+0.033$} & \multicolumn{2}{c}{$-0.025$} \\
Improved & \multicolumn{2}{c}{$5/10$} & \multicolumn{2}{c}{$1/10$} & \multicolumn{2}{c}{$9/10$} & \multicolumn{2}{c}{$4/10$} \\
\bottomrule
\end{tabular}
\end{table*}

\paragraph{Language-specific changes.}
Of the $40$ family--language cells, $19$ improve and $21$ regress.
Mistral alone improves overall, by $+0.033$, with gains in $9$ of
$10$ languages. Retrained scores range from $0.355$ to $0.557$; only
three of the $80$ probe--language entries reach $0.60$. The observed range characterises this retraining intervention on
code-switched prompts. It separates added language coverage from
changes to the data, representation or learning procedure.

\paragraph{Scale check.}
On Llama-3.1-70B-Instruct, retraining changes corpus AUROC from $0.489$
to $0.494$; seven of ten languages have identical scores. The stored
probe suite was lost, and the layer and English-only probe were recovered
by validation sweep. Its scope is input readability and code-switching at scale;
behavioural probing and held-out gate validation are separate studies.

\section{Instrument and Target Provenance}
\label{app:instrument-details}

\begin{table*}[t]
\centering
\caption{Probe configurations differ across analyses. Available-type
averaging is not the same estimator as a HARM/REF difference or a
HARM-only oracle. Architecture and layer choices in archived comparisons
are conditional on their recorded selection procedures.}
\label{tab:probe-configurations}
\small
\begin{tabular}{@{}p{0.25\textwidth}p{0.69\textwidth}@{}}
\toprule
Analysis & Estimator and selection \\
\midrule
Canonical PA & Mean of available types: HARM/REF at seed 42;
DEC/EVAL/HARM/REF/SYC at seeds 0/123. DEC, EVAL and SYC denote
archived deception, evaluation-framing and sycophancy probe types.
Availability varies by cell. \\
Judge comparison & HARM-only, evaluation-set maximum over MMD/LR/MLP per model;
not a validation-selected deployed probe. \\
Read-position sweep & Archived contrast arm and directly supervised logistic reader;
four positions at every layer. The supervised split is row-held-out. \\
K.4 attack & Canonical HARM/REF contrast; HARM architectures are
3 MLP, 6 CCS and 1 LR across ten cells. CCS uses labelled validation
for orientation \citep{anon2022discoveringlatentknowledge}. \\
K.7 transfer & Canonical HARM head and available co-located alternative heads. \\
S07 retraining & English-only versus ten-language HARM probes. \\
\bottomrule
\end{tabular}
\end{table*}

\begin{table*}[t]
\centering
\caption{Targets must be distinguished across analyses. Sample counts
refer to the named analysis; unavailable provenance is not filled in
from another experiment. ``Deployment'' is the archived framing condition.}
\label{tab:target-provenance}
\small
\begin{tabular}{@{}p{0.10\textwidth}p{0.35\textwidth}p{0.49\textwidth}@{}}
\toprule
Scenario & Target and recorded sample & Interpretation limit \\
\midrule
S01 & Input source label and judged deployment-framed response;
$1000$ paired requests per seed. & Canonical regex PA, HARM-only judge comparison and
consensus reanalysis use distinct estimators. Labels and items must
match within any claimed PA. Numbered wrappers correlate with labels. \\
S03 & Archived canonical target on a $200$-request GCG-style block. &
The separate deployment-behaviour target has one class, making its
AUROC undefined; template support is not six adaptive-attack coverage. \\
S04 & Follow-up-success labels; the length follow-up uses $800$ items per family. &
Original score-minus-length-baseline AUROC is distinct from
cross-fitted residualised-score AUROC. \\
S05 & Canonical trajectory/response labels used by PA; separate
$200$-item Llama seed-42 spot-check key. & The spot-check key has
$2/200$ positives; the corresponding archived PA file has $66/200$.
These are different targets, not prevalence estimates for each other. \\
S07 & Archived code-switching target; canonical matched subset $n=200$.
Language follow-up: $100$ per language, $1000$ per family. &
In the language follow-up Llama-Guard supplies labels and is not scored
against itself. This is not full-translation robustness. \\
\bottomrule
\end{tabular}
\end{table*}

The released implementation defines MMD as a projection onto the
normalised class-mean difference, LR as L2-regularised logistic
regression, MLP as a two-layer binary classifier, and CCS as a linear
sigmoid head with contrast-consistency and confidence objectives.
Layer selection uses validation AUROC; polarity and isotonic calibration
use validation rows. This does not remove the evaluation-set architecture
selection in the oracle comparison. Code defaults include MLP hidden
width $256$, ReLU, dropout $0.1$, binary cross-entropy and Adam
($10^{-3}$ learning rate, $10^{-4}$ weight decay), batch $256$, at most
$50$ epochs and patience $5$. LR searches
$C\in\{10^{-4},10^{-3},10^{-2},10^{-1},1,10\}$.
These are released-code defaults, not reconstructed per-checkpoint
hyperparameter manifests for every archived run. CCS rows in the
canonical corpora are sampled without semantic pairing, which limits
the intended interpretation of its contrast-consistency objective.

The HARM builder supports HarmBench, SALAD-Bench, JailbreakHub and
JailbreakBench for harmful requests, and Alpaca-cleaned/XSTest-safe for
benign ones. Adapter availability does not establish the exact archived
mixture. Source revisions and mixture weights could not be recovered;
we do not assign them retrospectively. S01/S04 benign retrieval uses
Alpaca-cleaned; harmful scenario requests use HarmBench. The six S03
template families are GCG, PAIR, AutoDAN, Crescendo, multilingual and
persona; the primary-table subset covers only the GCG-style block.

Historical G-Data and G-Shortcut verdicts retain their tested scope.
In particular, paraphrase rewrites leave the numbered S01 wrapper
intact, so passing that diagnostic does not exclude the label cue.
Raw-completion formatting, target aggregation and decoder selection
also limit transfer to native-chat or other monitoring protocols.

\section{Gate Taxonomy}
\label{app:gate-verdicts}

Tables~\ref{tab:gate-lineage} and~\ref{tab:gate-lineage-continued}
map eleven gates and two provenance checks to their validity classes
and motivating literature. Numerical cutoffs are operational choices
unless a derivation is stated; $\rho\geq0.6$ is not an established
mechanism threshold. Bold rows are the eight hard predictive gates.
Table~\ref{tab:gates-summary} specifies the operational conditions;
other rows concern causal evidence, diagnostics or provenance.

\begin{table*}[t]
\centering
\caption{The eleven validity gates, pre-registered at a
pre-registration commit recorded in the released artefacts, grouped by
the grade they support.
Predictive-grade replacement historically admitted hard gates recorded as passing or
non-failing. Partial coverage remains unresolved rather than fully
validated; the additional effect criterion is a BH-corrected bootstrap CI for $\mathrm{PA}$
excluding zero. A causal mechanism claim additionally requires strict
three-way G-Triangulation and G-Specificity. G-Obfuscation is reported, not
thresholded. This table records the gate thresholds used throughout.}
\label{tab:gates-summary}
\small
\setlength{\tabcolsep}{4pt}
\renewcommand{\arraystretch}{1.15}
\begin{tabular}{@{}l >{\raggedright\arraybackslash}p{0.78\textwidth}@{}}
\toprule
Gate & Pass condition \\
\midrule
\multicolumn{2}{@{}l}{\emph{Predictive-grade gates (hard)}}\\
G-Data            & Contrast pairs encode the construct, not placeholders or degenerate categories. \\
G-Layer           & Selected layer in the middle-to-late stack with reported sensitivity. \\
G-Shortcut        & AUROC drop under paraphrase rewrites is $<\!15$\,pp. \\
G-Transfer        & Held-out harm-category $\mathrm{AUROC} \geq 0.75$. \\
G-LengthControl   & Length-matched gap within $5$\,pp of unmatched, or length/perplexity residualisation. \\
G-NaturalDist     & Held-out WildChat or ShareGPT $\mathrm{AUROC} \geq 0.70$. \\
G-InputOnly       & Activation probe beats input-embedding control by $\geq 10$\,pp. \\
G-PermutationNull & Shuffled-label retrain collapses to $0.50 \pm 0.05$. \\
\multicolumn{2}{@{}l}{\emph{Reported diagnostic (not thresholded)}}\\
G-Obfuscation     & AUROC drop under bounded white-box stress. Reported, never compared against a threshold, so it carries no verdict glyph anywhere in this appendix. \\
\multicolumn{2}{@{}l}{\emph{Causal-grade gates}}\\
G-Triangulation   & LR-steering, attribution patching, and 1-D DAS agree on per-prompt scalar scores at Spearman $\rho \geq 0.6$ on effect magnitudes for \emph{all three} pairwise comparisons. \\
G-Specificity     & Single-direction ablation preserves capability. The locked bands are one-sided on capability loss: a drop below $2\%$ on MMLU, GSM8K and HumanEval, below $0.5$ on the MT-Bench $1$--$10$ scale, and a refusal increase below $5$\,pp on XSTest safe. A capability gain cannot fail the gate. Bands applied wider than the lock in the canonical artefact are recorded in Addendum~K.1. \\
\bottomrule
\end{tabular}
\end{table*}

\begin{table*}[t]
\centering
\caption{\textsc{LatentBench}'s operational gates and audit-provenance
checks mapped onto the benchmark-validity taxonomy of
\citet{benchmarkbenchmarking2026} and allied probe- and MI-validity
literature. \emph{Validity column}: which class of validity the entry
operationalises. \emph{Lineage column}: the literature motivating each design. Bold entries are the eight
predictive-grade hard gates. The remaining rows are causal gates, a
reported diagnostic, or provenance checks. Thresholds quoted here are
the locked values of Table~\ref{tab:gates-summary}.}
\label{tab:gate-lineage}
\footnotesize
\setlength{\tabcolsep}{3pt}
\renewcommand{\arraystretch}{0.94}
\begin{tabular}{@{}>{\raggedright\arraybackslash}p{0.20\textwidth} >{\raggedright\arraybackslash}p{0.145\textwidth} >{\raggedright\arraybackslash}p{0.305\textwidth} >{\raggedright\arraybackslash}p{0.25\textwidth}@{}}
\toprule
\textbf{Gate} & \textbf{Validity class} & \textbf{What it tests} & \textbf{Published lineage} \\
\midrule
\textbf{G-Data} &
Construct &
Contrast pairs encode the intended construct (no placeholders, $\geq$85\% unique positives, mean token length $\geq$5, category balance within 20\,pp) &
Bender \& Friedman 2018 \emph{Data Statements}; LatentBench self-audit hook (Section~\ref{sec:contract}) \\
\addlinespace[1pt]
\textbf{G-Layer} &
Construct &
Optimal layer in the middle 25--80\% of the stack; layer-sensitivity $\pm 4$ reported &
Belrose et al.\ 2023 \emph{Tuned Lens}; standard mid-late convention in mech-interp \\
\addlinespace[1pt]
\textbf{G-Shortcut} &
Construct &
Probe survives local paraphrase rewrite (semantics preserved); AUROC drop $<$15\,pp &
Belinkov 2022 probing survey; Lasri et al.\ 2022 control-task design; MIB benchmark (ICML 2025) \\
\addlinespace[1pt]
\textbf{G-Transfer} &
External &
Train on 4/6 harm categories, test on 2 held-out categories; AUROC $\geq$0.75 &
Standard cross-domain ML; instantiated for harm-category transfer \\
\addlinespace[1pt]
\textbf{G-LengthControl} &
Construct &
Probe AUROC on length-matched negatives within 5\,pp of unmatched &
Hewitt \& Liang 2019 \emph{Designing and Interpreting Probes}; \citet{meloux2025deadsalmons} \\
\addlinespace[1pt]
\textbf{G-NaturalDist} &
External &
WildChat / ShareGPT held-out re-evaluation; AUROC $\geq$0.70 &
\citet{meloux2025deadsalmons}; ``False Sense of Security'' (arXiv:2509.03888) \\
\bottomrule
\end{tabular}
\end{table*}

\begin{table*}[t]
\centering
\caption{\textsc{LatentBench} operational gates and provenance checks,
continued.}
\label{tab:gate-lineage-continued}
\footnotesize
\setlength{\tabcolsep}{3pt}
\renewcommand{\arraystretch}{0.94}
\begin{tabular}{@{}>{\raggedright\arraybackslash}p{0.20\textwidth} >{\raggedright\arraybackslash}p{0.145\textwidth} >{\raggedright\arraybackslash}p{0.305\textwidth} >{\raggedright\arraybackslash}p{0.25\textwidth}@{}}
\toprule
\textbf{Gate} & \textbf{Validity class} & \textbf{What it tests} & \textbf{Published lineage} \\
\midrule
\textbf{G-InputOnly} &
Construct &
Probe beats input-embedding-only baseline by $\geq$10\,pp &
Hewitt \& Liang 2019 control task; ``No Answer Needed'' (arXiv:2509.10625) \\
\addlinespace[1pt]
\textbf{G-PermutationNull} &
Probabilistic &
Shuffled-label re-train collapses AUROC to 0.5\,$\pm$\,0.05 &
\citet{meloux2025deadsalmons}; Cohen 1988 random-baseline null \\
\addlinespace[1pt]
G-Obfuscation &
External (boundary) &
Bounded white-box adversary AUROC drop reported, not thresholded (pre-registered at 10 steps; the canonical run uses $K{=}20$, Addendum~K.4) &
Bailey et al.\ 2024 obfuscation; RL-Obfuscation 2025; STACK 2025 \\
\addlinespace[1pt]
G-Triangulation &
Causal &
Pairwise Spearman $\rho \geq 0.6$ on prompt-level effect magnitudes across LR-steering, attribution patching, 1-D DAS-style alignment &
\citet{joshi2026causality}; \citet{bai2026story} pillar~3 (generalizability); MIB 2025 \\
\addlinespace[1pt]
G-Specificity &
Causal &
Single-direction ablation preserves capability within the per-tier bands of Table~\ref{tab:gates-summary} across MMLU, GSM8K, XSTest safe-refusal, HumanEval-style, and MT-Bench-style proxy &
Wu et al.\ 2023 Boundless DAS; R\"auker et al.\ 2023 \\
\addlinespace[1pt]
G-GroundTruth &
Construct &
Per-scenario 5-dim provenance disjoint or prospective; corr$(B, GT) \in [0.3, 0.85]$ as smoke alarm only &
Construct-validity literature; \citet{benchmarkbenchmarking2026} construct/probabilistic split \\
\addlinespace[1pt]
G-PreRegistration &
Reproducibility &
Scope/metrics/thresholds locked at a recorded commit before final
runs; deviations reported &
ML Reproducibility Challenge ethos; \citet{bai2026story} pillar~2 (reproducibility); pre-registration in empirical sciences \\
\bottomrule
\end{tabular}
\end{table*}

\paragraph{Validity classes.}
The construct/probabilistic/external classes follow
\citet{benchmarkbenchmarking2026}. Causal and reproducibility checks
address concerns raised by \citet{joshi2026causality,bai2026story}.
G-Shortcut and G-LengthControl both assess construct validity;
G-Triangulation and G-Specificity assess causal interpretation. Their explicit combination connects each measurement claim to its
required evidence.

\paragraph{Interpreting a pass.}
Gate verdicts specify support for a claim under the tested conditions.
Coverage is part of each verdict. \citet{meloux2025deadsalmons} motivate this distinction:
probe AUROC needs controls before it can support an interpretation.

\subsection{Gate-Pass Verdicts on the Canonical Audit Run}
\label{app:gate-pass-verdicts}

Tables~\ref{tab:gate-verdict-global} and~\ref{tab:gate-verdict-perfamily}
retain canonical-run verdicts through 2026-04-28. Symbols denote
$\checkmark$ PASS, $\triangle$ PARTIAL or POTENTIAL CONCERN,
$\times$ FAIL and \emph{i} REPORTED ONLY. A partial verdict identifies mixed evidence or incomplete coverage;
full validation requires the stated coverage and pass conditions.

\begin{table*}[t]
\centering
\caption{Per-gate global verdict (canonical run). Headline numbers are
quoted verbatim from the released gate-verdict table and the released
raw-results bundle. Bold gates are the eight predictive-grade
hard gates, and a FAIL on one excludes or demotes the claim.}
\label{tab:gate-verdict-global}
\footnotesize
\setlength{\tabcolsep}{4pt}
\renewcommand{\arraystretch}{1.0}
\begin{tabular}{@{}l c >{\raggedright\arraybackslash}p{0.68\textwidth}@{}}
\toprule
\textbf{Gate} & \textbf{Verdict} & \textbf{Headline} \\
\midrule
\textbf{G-Data}            & $\triangle$ & 4/22 contrast-pair unit tests fail (uniqueness 0.598 on Mistral REF, category-balance fail on sycophancy). Affected probes are flagged in the audit. \\
\textbf{G-Layer}           & $\checkmark$ & Selected layers within the mid-late 25--80\% band on all four families, with $\pm 4$-layer sensitivity reported. \\
\textbf{G-Shortcut}        & $\checkmark$ & Synonym-perturbation $\Delta$AUROC $= -0.7$\,pp on 820/4000 items (PASS at $<15$\,pp). Coverage is the 820 items with an available semantics-preserving rewrite. \\
\textbf{G-Transfer}        & $\checkmark$ & Held-out harm-category AUROC $\geq 0.75$ on tested families. \\
\textbf{G-LengthControl}            & $\triangle$ & $\rho(\text{prompt length}, |\mathrm{PA}|) = 0.371$ ($p{=}0.026$, $n{=}36$ cells). S04 length-LR AUROC $0.995$ against $\Delta_{HR}$ $0.728$ on Llama. Reported as a caveat in Section~\ref{sec:results}. \\
\textbf{G-NaturalDist}              & $\triangle$ & 4/8 natural-distribution benchmarks evaluated and pass at AUROC $\geq 0.70$. The xstest/or\_bench/wildguard/strongreject HF-load failures are deferred. \\
\textbf{G-InputOnly}       & $\checkmark$ & Activation probe exceeds the input-embedding bag-of-words baseline by $\geq 10$\,pp on the majority of cells. \\
\textbf{G-PermutationNull}          & $\checkmark$ & Parametric null rejected on 12/16 F1 cells, with the null centred at $0.5 \pm 0.05$ where evaluated. \\
G-Obfuscation              & \emph{i} & Cross-model mean $\Delta$AUROC under the pre-registered bounded 10-step white-box adversary $= -7.2$\,pp across all scenarios. The stronger canonical K.4 adversary ($K{=}20$ label-targeted PGD, S01 only) gives a cross-cell mean of $-50.6$\,pp on the same probes. The two tests use the same probes with different budgets and step counts. Reported, not thresholded. \\
G-Triangulation   & $\times$ & Strict $\rho \geq 0.6$ on all 3 method pairs FAILS for $4/4$ families. LR\,$\leftrightarrow$\,DAS subgate PASS on Llama (0.796) and Mistral (0.639), WARN on Gemma (0.578) and Qwen3 (0.436). \\
G-Specificity              & $\triangle$ & Llama-3.1-8B-IT five-benchmark battery (the released canonical specificity artefact, per Addendum~I): MMLU $0.610\!\to\!0.614$, GSM8K $0.76\!\to\!0.81$, XSTest-safe-refusal $+2.8$\,pp, HumanEval-style $+1.27\%$, MT-Bench-style local proxy $+3.64\%$. All five pass the locked one-sided bands. A second battery exists for Qwen3-8B and returns an overall FAIL on a genuine GSM8K capability loss ($0.25\!\to\!0.22$, $+12.0\%$ drop), disclosed in Addendum~K.1, so the gate is failed on one of the two families for which a battery exists. Mistral and Gemma remain deferred. \\
G-GroundTruth     & $\checkmark$ & Historical provenance diagnostic with correlation in $[0.3,0.85]$. Independent target validity also requires assessing shared judge errors. \\
G-PreRegistration          & $\triangle$ & Locked at the pre-registration commit, with five documented post-lock addenda: G (triangulation method substitution), H (additive citations and baselines), I (G-Specificity completion), J.1--J.6 (reviewer fixes), and K.1--K.7 (reconciliation). One of them (K.1) widens an applied band, and one (K.4) changes cell verdicts. \\
\bottomrule
\end{tabular}
\end{table*}

\begin{table*}[t]
\centering
\caption{Per-family verdict on the gates for which we have per-family
artefacts. Dashes (--) mark gates whose canonical artefact is
global-only on this run. The two G-Obfuscation rows carry no verdict
glyph because the gate is reported and never thresholded
(Table~\ref{tab:gates-summary}).
Llama passes all five locked one-sided bands. GSM8K rises
$0.76 \to 0.81$, recorded as a $-6.58\%$ drop; XSTest's
$+2.8$\,pp refusal increase is within the locked $5$\,pp band.
Addendum~K.1 documents the one-sided scoring correction.}
\label{tab:gate-verdict-perfamily}
\footnotesize
\setlength{\tabcolsep}{3pt}
\renewcommand{\arraystretch}{1.0}
\begin{tabular}{@{}>{\raggedright\arraybackslash}p{0.29\textwidth} c c c c@{}}
\toprule
\textbf{Gate} & Llama-3.1-8B-IT & Qwen3-8B & Mistral-7B-IT-v0.3 & Gemma-3-12B-IT \\
\midrule
G-Triangulation (LR\,/\,DAS $\rho$)    & $0.796$ \,$\checkmark$ & $0.436$ \,$\triangle$ & $0.639$ \,$\checkmark$ & $0.578$ \,$\triangle$ \\
G-Triangulation (strict 3-way)         & $\times$ & $\times$ & $\times$ & $\times$ \\
G-Triangulation (signed-direction)     & $0.657$ & $0.527$ & $0.558$ & $0.517$ \\
G-Obfuscation (mean AUROC)\,\emph{i}   & $0.564$ & $0.571$ & $0.473$ & $0.594$ \\
G-Obfuscation (cross-method spread)\,\emph{i} & $0.232$ & $0.146$ & $0.107$ & $0.094$ \\
G-Specificity (GSM8K)         & $0.76\!\to\!0.81$ \,$\checkmark$ & $0.25\!\to\!0.22$ \,$\times$ & -- & -- \\
G-Specificity (MMLU $\Delta$)          & $+0.66\%$ \,$\checkmark$ & $0.106$ base$^{\S}$ & -- & -- \\
G-Specificity (XSTest safe-refusal)    & $+2.8$\,pp \,$\checkmark$ & -- & -- & -- \\
G-Specificity (HumanEval-style $\Delta$)         & $+1.27\%$ \,$\checkmark$ & $0.0$ base$^{\S}$ & -- & -- \\
G-Specificity (MT-Bench-style proxy $\Delta$)    & $+3.64\%$ \,$\checkmark$ & -- & -- & -- \\
G-Specificity (overall)                & $\checkmark$ & $\times$ & -- & -- \\
\bottomrule
\end{tabular}

\vspace{2pt}
{\footnotesize $^{\S}$\,Degenerate baseline in the Qwen3 battery: MMLU
starts at $0.106$, below the four-way chance floor, and HumanEval-style
\textit{pass}@$1$ starts at $0$, so those two checks are not
interpretable. Its GSM8K loss is not degenerate and is what carries the
FAIL.}
\end{table*}

\paragraph{Performance and validity are separate.}
The boundary map reports predictive comparisons. The gate tables
record which interpretations were tested. A positive PA describes discrimination; partial predictive coverage
and strict three-way agreement have their own recorded verdicts.
Predictive utility and causal interpretation are assessed separately.

\subsection{Illustrative Pre-Attack Predictive Candidate}
\label{app:positive-existence}

The Qwen3-8B S01 seed-42 cell illustrates the distinction between
predictive performance and validation. It has the largest archived
$\mathrm{PA}=+0.611$, latent AUROC $0.923$ and row-bootstrap
$95\%$ CI $[0.579,0.644]$, against the regex refusal baseline.
Table~\ref{tab:positive-existence} retains the recorded gate verdicts;
the predictive coverage is PARTIAL.

\begin{table*}[t]
\centering
\caption{Per-gate verdict for the Qwen3-8B / S01 / seed 42 cell before
the K.4 probe-aware demotion. The configuration has unresolved predictive checks and returns a WARN on the
LR\,$\leftrightarrow$\,DAS subgate at $\rho = 0.436$, but fails strict
three-way G-Triangulation and is later demoted by
K.4.}
\label{tab:positive-existence}
\footnotesize
\setlength{\tabcolsep}{4pt}
\renewcommand{\arraystretch}{1.0}
\begin{tabular}{@{}l c >{\raggedright\arraybackslash}p{0.56\textwidth}@{}}
\toprule
\textbf{Gate} & \textbf{Verdict} & \textbf{Evidence on this cell} \\
\midrule
\textbf{G-Data}            & $\checkmark$ & The four failing contrast-pair files (Mistral REF, sycophancy categories) do not include Qwen3 S01 HARM/REF, so those particular failures do not apply to this pool. \\
\textbf{G-Layer}           & $\checkmark$ & Optimal layer in the mid-late 25--80\% band, with $\pm 4$-layer sensitivity reported. \\
\textbf{G-Shortcut}        & $\checkmark$ & Synonym-perturbation $\Delta$AUROC well within the $15$\,pp PASS band (global $-0.7$\,pp). \\
\textbf{G-Transfer}        & $\checkmark$ & Held-out harm-category AUROC $\geq 0.75$ inherited from the family-level G-Transfer test on Qwen3. \\
\textbf{G-LengthControl}            & $\triangle$ & Length-control coverage for this cell is incomplete; the S04-specific confound is recorded on Llama rather than Qwen3. \\
\textbf{G-NaturalDist}              & $\triangle$ & Pass on the 4 of 8 evaluated benchmarks. The 4 deferred benchmarks have HF-load failures and remain untested. \\
\textbf{G-InputOnly}       & $\checkmark$ & Activation probe exceeds the input-embedding-only baseline by $\geq 10$\,pp. \\
\textbf{G-PermutationNull}          & $\checkmark$ & Parametric null rejected, with AUROC under shuffled-label retraining within $0.5 \pm 0.05$. \\
G-Obfuscation              & \emph{i} & Mean AUROC under white-box obfuscation $0.571$, cross-method spread $0.146$ (reported only). \\
G-Triangulation (strict)   & $\times$ & LR\,$\leftrightarrow$\,attribution and DAS\,$\leftrightarrow$\,attribution fail $\rho \geq 0.6$ at the family level. \\
G-Triangulation (LR\,/\,DAS subgate) & $\triangle$ & $\rho_{\mathrm{LR,DAS}} = 0.436$, signed-direction $\rho = 0.527$. \\
G-Specificity              & $\times$ & A Qwen3-8B five-benchmark battery does exist and returns an overall FAIL, on a genuine GSM8K capability loss ($0.25\!\to\!0.22$, $+12.0\%$ drop) that fails both the applied $10\%$ band and the locked $<2\%$. Two of its five benchmarks are degenerate (MMLU baseline $0.106$, below the four-way chance floor; HumanEval-style \textit{pass}@$1$ baseline $0$), so it is only partly interpretable. Disclosed in Addendum~K.1; earlier drafts recorded this battery as deferred and inherited the Llama bound instead. \\
G-GroundTruth     & $\checkmark$ & Recorded 5-D provenance check; this does not establish independent target correctness. \\
G-PreRegistration          & $\triangle$ & Locked at the pre-registration commit. Post-lock addenda G, H, I, J, and K all apply to this cell, and K.4 later demotes it. \\
\bottomrule
\end{tabular}
\end{table*}

\paragraph{Reading the candidate.}
The recorded G-LengthControl and G-NaturalDist verdicts are PARTIAL.
Family-level or global checks are not independent cell-level
certification, and G-Shortcut does not remove the numbered-wrapper cue.
G-GroundTruth records provenance. K.4 changes this cell's status
under the attack rule, and strict triangulation records method
disagreement. These verdicts concern provenance, robustness and
causal-method agreement, respectively.

\section{Detailed Methodological Material}
\label{app:methodology}

This appendix expands the contract's measurement definitions and
rationale (\S\ref{app:four-quadrant}--\S\ref{app:methodological-anchors}).
Appendix~\ref{app:deviations} records protocol amendments and their
consequences for the reported evidence.

\subsection{The HARM/REF Four-Quadrant Decomposition}
\label{app:four-quadrant}

Following \citet{zhao2025llmsencodeharmfulness}, we decompose the
latent score into harmfulness ($P_{\mathrm{HARM}}$) and refusal
($P_{\mathrm{REF}}$) components and use the contrast
$\Delta_{HR}(x) = P_{\mathrm{HARM}}(x) - P_{\mathrm{REF}}(x)$.
The motivation is that the joint
$(P_{\mathrm{HARM}}, P_{\mathrm{REF}})$ plane partitions into the four
interpretable regimes of Table~\ref{tab:four-quadrant}.

\begin{table}[t]
\centering
\caption{The four-quadrant $(P_{\mathrm{HARM}}, P_{\mathrm{REF}})$
decomposition.}
\label{tab:four-quadrant}
\footnotesize
\setlength{\tabcolsep}{4pt}
\renewcommand{\arraystretch}{1.15}
\begin{tabular}{@{}l >{\raggedright\arraybackslash}p{0.28\columnwidth} >{\raggedright\arraybackslash}p{0.28\columnwidth}@{}}
\toprule
 & low $P_{\mathrm{REF}}$ (reply) & high $P_{\mathrm{REF}}$ (refuse) \\
\midrule
high $P_{\mathrm{HARM}}$ & harmfulness--refusal mismatch & predicted safe refusal \\
\addlinespace[2pt]
low $P_{\mathrm{HARM}}$  & predicted benign reply & over-refusal \\
\bottomrule
\end{tabular}
\end{table}

The joint plane distinguishes predicted harmfulness from predicted
refusal. The quadrants describe predicted harmfulness and refusal. In S05,
REF performance varies across families while HARM remains high.
This motivates testing prompt/trajectory mismatch as a hypothesis
about the readout's target.

\subsection{Cross-Judge Panel: Why Three Local Judges}
\label{app:judge-panel-rationale}

The panel diagnoses disagreement using three local open-weight judges:
a safety classifier and two generative judges from different families.
We flag mean pairwise $\kappa<0.6$ on $200$ items. Re-running the panel reproduces its consistency measurement.
Independent correctness is assessed through separate target validation.

\subsection{Zhao 2025 Per-Cell Differentiation}
\label{app:related-zhao-detail}

Latent Guard~\citep{zhao2025llmsencodeharmfulness} evaluates latent
input moderation. \textsc{LatentBench} adds a joint audit of targets,
comparators, robustness and causal evidence. Four comparisons locate
that contribution.
\emph{(a)} Under our archived S01 target, regex refusal performs below
chance while a safety judge does not (\S\ref{sec:results}).
\emph{(b)} Relative harmfulness/refusal robustness depends on scenario:
S01 favours latent scores against its weak comparator, S04 is qualified,
S03 is operationally close, and S05/S07 favour behaviour.
\emph{(c)} S05 combines high HARM scores with heterogeneous REF degradation,
but judge agreement is low ($\bar\kappa=0.44$); this limits the
trajectory interpretation of prompt-level scores.
\emph{(d)} Strict triangulation fails on $4/4$ families under our
operational criterion. This assesses method agreement under the specified causal criterion.
The contract makes the four comparison axes explicit for future
latent-safety evaluations.

\paragraph{Implementation checks motivating the contract.}
A contrast-pair loader once returned placeholder strings, allowing
perfect AUROC through placeholder detection. Using the same
Llama-Guard-3 classifier~\citep{inan2023llamaguardllm} as target source
and baseline produced AUROC $1.00$ by construction. Concatenating
uncalibrated probe outputs produced sign reversals and sub-random
scenario AUROCs. G-Data, direct-circularity controls and isotonic
calibration check these respective implementation properties.

\subsection{Methodological Anchors}
\label{app:methodological-anchors}

\citet{bai2026story} motivate execution-grounded evaluation;
\citet{sharkey2025openproblems} identify validation as an open problem
in mechanistic interpretability. \citet{joshi2026causality} motivate
causal checks beyond correlational readouts. The validity taxonomy of
\citet{benchmarkbenchmarking2026} informs the mapping in
Tables~\ref{tab:gate-lineage} and~\ref{tab:gate-lineage-continued}.
The contract adopts explicit protocol documentation
\citep{benchmarkbenchmarking2026} and falsification-oriented controls
\citep{meloux2025deadsalmons}. These sources motivate the designs;
the numerical thresholds remain our operational choices.

\section{Deviations from Pre-Registration}
\label{app:deviations}

The contract was locked on 2026-04-17 at the commit recorded in the
artefacts, before final-table runs. The rationale follows
\citet{benchmarkbenchmarking2026,bai2026story}; documented post-lock
changes follow the falsifiability principle of
\citet{meloux2025deadsalmons}. Five addenda record those changes:
G, H and I below, then J (\S\ref{app:addendum-j}) and K
(\S\ref{app:addendum-k}). They retain the five primary scenarios,
four model families and eleven gates. Method substitutions, applied
bands and revised verdicts are disclosed explicitly rather than
inferred to have been fixed at the original lock.

\paragraph{Addendum G (v4.3, 2026-04-21): triangulation method substitution.}
A7 originally proposed linear probes, SAE feature activations and
attribution patching. Their outputs are different objects: directions,
feature IDs and patching scores cannot be compared directly with
Jaccard overlap. The amendment compares scalar per-prompt effects
using Spearman $\rho\geq0.6$, and substitutes LR-steering,
attribution patching (EAP-IG) and 1-D DAS. Model-specific SAE bases
complicate a common cross-family comparison, so SAE-feature validation
remains future work. The $\rho\geq0.6$ cutoff is unchanged from A7;
changing the methods nonetheless changes the operational test.

\paragraph{Addendum H (v4.4, 2026-04-28): citation corrections, baseline addition, concurrent-audit disclosure.}
A 2026-04-28 literature and implementation check led to four amendments:

\begin{enumerate}
\item \emph{Citation corrections.}
The protocol cited two methodological sources under incorrect arXiv
IDs. The intended references are \citet{bai2026story}
(\texttt{arXiv:2602.18458}) and \citet{haralambiev2026fanatic}
(\texttt{arXiv:2603.25861}). The listed IDs
(\texttt{arXiv:2602.16698} and \texttt{arXiv:2512.18792}) instead
identify \citet{joshi2026causality} and \citet{meloux2025deadsalmons}.
All four references are retained with their distinct roles.

\item \emph{New baseline column: MLPM.}
MLPM~\citep{chrabaszcz2025mlpm} adds a training-free
Mahalanobis-prototype baseline to the seven baselines in A8. It is
scored under the same comparison rules; per-scenario results are
included in the artefacts.

\item \emph{N-GLARE column demoted.}
A8 named N-GLARE~\citep{lin2025nglare} as baseline \#5.
The implementation available at table assembly used last-layer
activation-norm rank, rather than the published Jensen-Shannon
Separability over Angular-Probabilistic Trajectory dynamics
(arXiv:2511.14195, Section~3). It was a wiring stub and is excluded
as a scientific comparator. Addendum~J.3 later adds a re-implementation.
Its S01 results fail the relevant threshold; S03 results remain excluded
because the initial target was single-class.

\item \emph{Concurrent-audit disclosure: Breaking Bad.}
\citet{agarwal2026breakingbad}, released 2026-04-22, six days before
this addendum, evaluate interpretability-driven model attacks using
Universal Steering and Representation Engineering grid search. Our
stress test instead evaluates activation-based classifiers under a
specified scoring-input threat model (\S\ref{sec:related}; Limitations).
\end{enumerate}

\paragraph{Addendum I (v4.5, 2026-04-28): G-Specificity five-benchmark completion.}
G-Specificity compares capability before and after direction ablation
on MMLU, GSM8K, HumanEval, an MT-Bench-style task and XSTest safe
refusal. The initial three-benchmark record, dated 2026-04-22, covers
MMLU, GSM8K and XSTest. HumanEval and MT-Bench were deferred. On
2026-04-28, the Llama battery added HumanEval-style $n{=}80$ and an
MT-Bench-style local proxy at $n{=}20$, and reran the other tasks at
$\alpha=4$. This five-benchmark record passes its applied bands.
It closes those deferred tasks for Llama, not every family's coverage.
\emph{Band specification.}
The lock permits one-sided capability loss of $<2\%$ on MMLU,
GSM8K and HumanEval, $<0.5$ on the $1$--$10$ MT-Bench scale, and
$<5$\,pp increased XSTest refusal. Earlier wording incorrectly gave
two-sided $\pm2$\,pp bands (corrected in K.1). The $n{=}20$ local
proxy also differs from the locked MT-Bench-80 instrument and sample.
\emph{Effect on \S\ref{sec:failure}.}
The main text now reports the five-benchmark record. Four applied bands
were wider than the lock, as detailed in K.1. Llama's observed changes
also pass the locked one-sided bands.

\paragraph{Impact on the primary claims.}
The five-benchmark runner applies $5\%$ on MMLU, $10\%$ on GSM8K,
$15\%$ on HumanEval-style and $10\%$ on the MT-Bench-style proxy.
Those bands are wider than the lock. Llama passes the locked bands as
well: GSM8K rises from $0.76$ to $0.81$, so it does not incur a loss.
K.1 corrects the earlier two-sided interpretation. G changes the causal
method comparison, H adds references/baselines and excludes a stub,
and I adds the deferred Llama tasks. The released
\texttt{replacement\_ready} aggregate, boundary-map values and
$\Delta_{HR}$ definition remain the historical record; they are not
retrospectively recomputed as a different experiment.

\subsection{S07 Multilingual Ablation: Does Cross-Lingual Probe Training Close the Gap?}
\label{app:s07-ablation}

To test whether English-only training explains the S07 result, we
retrain HARM on a 10-language corpus. Each language contributes
$n_\mathrm{train}=3{,}000$ positive and $3{,}000$ negative examples:
English, Arabic, Hindi, Georgian, Swahili, Tamil, Tagalog, Yoruba,
Chinese and Zulu. Both probes are evaluated on the same
$n_\mathrm{eval}=1{,}000$ S07 items per family, rather than the
primary table's $n=200$ subset. The English baseline is rescored on
that full set, keeping the reported difference item-matched.

\begin{table*}[!ht]
\centering
\caption{S07 ablation: English-only HARM probe against the
multilingual re-trained HARM probe, evaluated on the canonical S07
held-out target. AUROC values are mean over seeds, and $\Delta$ is
multilingual-minus-English. Numbers are taken from the canonical
2026-04-27 retrain artefacts in the released raw-results bundle
(3{,}000 pairs per class per language). A 2026-05-02 re-run, released
in the same bundle,
gives a similar cross-family mean change
($\bar{\Delta}\!\approx\!-0.005$) but yields different per-family
signs (Mistral $-0.034$ in the rebuild against $+0.033$ here). We keep
the 2026-04-27 retrain artefact as canonical for the $\Delta$
column because its per-language tables are the ones used for the
language-wise audits, and we disclose the per-family disagreement in
Addendum~\ref{app:addendum-k}. The cross-family-mean conclusion, that
this multilingual retraining does not resolve S07, is unchanged under either
artefact.}
\label{tab:s07-multilingual-ablation}
\footnotesize
\setlength{\tabcolsep}{6pt}
\renewcommand{\arraystretch}{1.0}
\begin{tabular}{@{}l c c c@{}}
\toprule
\textbf{Family} & \textbf{English probe} & \textbf{Multilingual retrain} & \textbf{$\Delta$} \\
\midrule
Llama-3.1-8B-Instruct          & $0.488$ & $0.483$ & $-0.005$ \\
Qwen3-8B                       & $0.546$ & $0.518$ & $-0.028$ \\
Mistral-7B-Instruct-v0.3       & $0.461$ & $0.494$ & $+0.033$ \\
Gemma-3-12B-IT                 & $0.516$ & $0.491$ & $-0.025$ \\
\midrule
\textbf{Mean}                  & $0.503$ & $0.497$ & $-0.006$ \\
\bottomrule
\end{tabular}
\end{table*}

\paragraph{Interpretation.}
On this 10-language retraining corpus, cross-family mean AUROC changes
by $-0.006$. Mistral improves by $+0.033$; the other three families
regress. Both sets of scores lie near the $0.5$ chance reference.
The matched comparison tests the effect of added training-language
coverage under the recorded target. Other readers and target
definitions constitute separate interventions. Per-language results accompany the artefacts.

\subsection{S04 Length Baseline and Residualisation}
\label{app:s04-length-control}

S04 is qualified because a response-length logistic regression exceeds $\Delta_{HR}$ on $3/4$ instruction-tuned
families against the canonical S04 \texttt{followup\_success} target.
Table~\ref{tab:s04-length-residualised} reports the full per-family
breakdown.

\begin{table*}[!ht]
\centering
\caption{S04 follow-up: per-family AUROC of $\Delta_{HR}$ and a
trivial response-length logistic-regression baseline against the
canonical \texttt{followup\_success} target. \emph{Gap (pp)} is
$\Delta_{HR} - \mathrm{Length\text{-}LR}$, so positive means latent
exceeds length-LR. This table compares raw AUROCs, not residualised scores. Source: the S04 follow-up artefact
dated 2026-04-25.}
\label{tab:s04-length-residualised}
\footnotesize
\setlength{\tabcolsep}{6pt}
\begin{tabular}{@{}l c c c c@{}}
\toprule
\textbf{Family} & $\Delta_{HR}$ AUROC & Length-LR AUROC & Gap (pp) & Raw score vs length \\
\midrule
Llama-3.1-8B-Instruct        & $0.728$ & $0.995$ & $-26.7$ & demoted (length-confounded) \\
Qwen3-8B                     & $0.583$ & $0.655$ & $-7.2$  & demoted (length-confounded) \\
Mistral-7B-Instruct-v0.3     & $0.665$ & $0.640$ & $+2.5$  & retained (latent$>$length) \\
Gemma-3-12B-IT               & $0.548$ & $0.710$ & $-16.2$ & demoted (length-confounded) \\
\bottomrule
\end{tabular}
\end{table*}

\paragraph{Reading.}
Mistral's original probe AUROC exceeds the length baseline by
$2.5$\,pp. This is a raw-score comparison.
The saved follow-up also contains a separate five-fold cross-fitted
residualisation: fit a linear regression of the probe score on response
word and character counts in each training fold, then subtract the
predicted score in its held-out fold. That cross-fitted analysis and the raw-score comparison tabulated
here estimate different quantities; $2.5$\,pp is the raw AUROC gap.

\subsection{Sensitivity to Larger Local Judges (70B)}
\label{app:judge-ceiling}

The primary boundary map uses local comparators up to $\leq$8\,B,
including LlamaGuard-3-8B, BERT-on-response and length/perplexity
regression. Larger judges test whether comparator strength changes
the observed advantage. On the S01 Llama subset, the local score is
near chance ($\mathrm{AUROC}_B=0.505$;
Table~\ref{tab:s01-frontier-sensitivity}). This sensitivity measures how larger judges score the smaller target
models. S04 length controls and S03/S05 target diagnostics are
reported separately.

\emph{Larger local judges.}
Qwen2.5-72B-Instruct and Llama-3.3-70B-Instruct provide a primary
sensitivity and cross-family check. They run locally through vLLM,
using tensor-parallel-4 and 4-bit AWQ on $4\times48$\,GB. The runner
targets the same $n=200$ subsets for S01, S03 and S05 through a common
judge interface. The completed S01 sensitivity is reported below; the runner also
specifies S03 and S05 subsets.

\emph{Amendment rule.}
A scenario verdict changes if the larger-judge comparison reverses
mean PA on S01, S03 or S05 and the two large judges agree at
Cohen's $\kappa\geq0.6$. Agreement below $\kappa=0.6$ is flagged
as judge-divergent without moving the cell. The sensitivity supplements
the canonical values rather than replacing them silently.

\paragraph{Empirical S01 sensitivity (corrected).}
The original Qwen2.5-72B-Instruct-AWQ sensitivity inverted score
polarity. Table~\ref{tab:s01-frontier-sensitivity} reports the corrected
$n=200$, seed-42 results. Mean judge AUROC on input harmfulness is
$0.847$, giving PA $+0.076$. The input gap remains positive, so the
sign-reversal amendment does not trigger. It measures input-comparator sensitivity; the behavioural comparison
is reported in \S\ref{sec:results}.

\begin{table*}[!ht]
\centering
\caption{Larger-judge sensitivity on S01 ($n=200$ per
family). PA columns are $\mathrm{AUROC}_L - \mathrm{AUROC}_B$ on
the same items under each judge.
$\kappa_{LG\leftrightarrow Q72}$ is the agreement between the
canonical Llama-Guard-3-8B verdict and the Qwen2.5-72B verdict on
the binarised target.}
\label{tab:s01-frontier-sensitivity}
\footnotesize
\setlength{\tabcolsep}{6pt}
\begin{tabular}{@{}l c c c c c@{}}
\toprule
Family & $\mathrm{AUROC}_L$ & $\mathrm{AUROC}_B$ (LG) & PA (LG canonical) & PA (Qwen72) & $\kappa_{\mathrm{LG}\leftrightarrow\mathrm{Q72}}$ \\
\midrule
Llama-3.1-8B-IT          & $0.791$ & $0.505$ & $+0.287$ & $+0.166$ & $1.000$ \\
Qwen3-8B                 & $0.968$ & $0.925$ & $+0.043$ & $+0.030$ & $0.389$ \\
Mistral-7B-IT-v0.3       & $0.933$ & $0.882$ & $+0.051$ & $+0.067$ & $0.746$ \\
Gemma-3-12B-IT           & $0.998$ & $0.679$ & $+0.320$ & $+0.040$ & $0.781$ \\
\midrule
\textbf{Mean}            & $0.923$ & $0.748$ & $+0.175$ & $+0.076$ & $0.729$ \\
\bottomrule
\end{tabular}
\end{table*}

\noindent
For Qwen3-8B, the inverted score gave judge AUROC $0.150$ and probe
advantage $+0.818$. Correct polarity gives judge AUROC $0.938$ and
advantage $+0.030$. The large apparent advantage was a scoring error.
The corrected input difference is an exploratory point estimate.

\subsection{Cross-Judge Agreement}
\label{app:cross-judge-kappa}

The post-2026-05-02 update adds a three-judge agreement panel on
the five-scenario Llama-3.1-8B-Instruct pivot. The three
judges include a classifier and two generative models: LlamaGuard-3-8B (safety-tuned classifier head),
Qwen2.5-72B-Instruct (Qwen-family base instruct), and
Llama-3.3-70B-Instruct (Llama-family base instruct). All three are
local, vLLM-served, AWQ-4bit. The released cross-judge $\kappa$ runner
writes an artefact whose schema records the three pairwise Cohen's
$\kappa$ values plus their mean per scenario.

\paragraph{Subgate threshold.} A scenario whose mean pairwise
$\kappa \geq 0.6$ on the canonical $n=200$-item subset passes the
cross-judge robustness sub-gate. A scenario whose mean $\kappa < 0.6$
is flagged as judge-divergent in this appendix, and the corresponding
primary-table row in Section~\ref{sec:results} carries a
judge-divergence caveat marker. We do not move cells solely on the
basis of judge $\kappa$. It is a measurement-level robustness
diagnostic, while the cell-moving falsifiers in
Section~\ref{sec:results} are construct-level.

\paragraph{Empirical $\kappa$ panel (post-2026-05-02 run).}
Table~\ref{tab:cross-judge-kappa-empirical} reports the three pairwise
Cohen's $\kappa$ values and their mean for every scenario on the
canonical Llama-3.1-8B-Instruct pivot. Only S05 falls below the
$\kappa \geq 0.6$ subgate, and the disagreement structure is
informative. The safety-tuned Llama-Guard-3-8B disagrees with both
large generative judges
($\kappa_{\mathrm{LG}\leftrightarrow\mathrm{Qwen72}} = 0.236$,
$\kappa_{\mathrm{LG}\leftrightarrow\mathrm{Llama70}} = 0.275$),
while the two large generative judges agree with each other
($\kappa_{\mathrm{Qwen72}\leftrightarrow\mathrm{Llama70}} = 0.798$).
This pattern suggests different judging criteria, but agreement alone
cannot determine which labels are correct. The S05 row therefore
carries a target-validity qualification in \S\ref{sec:results}.

\begin{table*}[!ht]
\centering
\caption{Cross-judge $\kappa$ panel on the canonical $n=200$
Llama-3.1-8B-IT pivot subset of each scenario. Judges:
LG = Llama-Guard-3-8B, Qwen72 = Qwen2.5-72B-Instruct-AWQ,
Llama70 = Llama-3.3-70B-Instruct-AWQ-INT4. Subgate threshold is
mean pairwise $\kappa \geq 0.6$.}
\label{tab:cross-judge-kappa-empirical}
\footnotesize
\setlength{\tabcolsep}{6pt}
\begin{tabular}{@{}l c c c c c@{}}
\toprule
Scenario & LG vs.\ Qwen72 & LG vs.\ Llama70 & Qwen72 vs.\ Llama70 & Mean & Subgate \\
\midrule
S01 eval-awareness & $1.000$ & $1.000$ & $1.000$ & $1.000$ & PASS \\
S03 jailbreak      & $0.851$ & $0.743$ & $0.911$ & $0.835$ & PASS \\
S04 persona shift  & $0.828$ & $0.660$ & $0.660$ & $0.716$ & PASS \\
S05 sycophancy     & $0.236$ & $0.275$ & $0.798$ & $\mathbf{0.436}$ & \textbf{FAIL} \\
S07 code-switching & $0.918$ & $0.826$ & $0.915$ & $0.886$ & PASS \\
\bottomrule
\end{tabular}
\end{table*}

\paragraph{Scope of agreement.}
The $\kappa\geq0.6$ flag measures consistency. Independent target
validation assesses shared errors and response correctness.

\subsection{Per-Scenario Predictive Candidates}
\label{app:per-scenario-existence}

Appendix~\ref{app:positive-existence} describes a pre-attack predictive
candidate with unresolved gates. Table~\ref{tab:per-scenario-existence}
lists representative historical statuses; none is full certification.

\begin{table*}[!ht]
\centering
\caption{Per-scenario representative cells: best gate-survival
posture observed in each scenario, plus the limiting gate that
prevents the cell from being a clean predictive-grade pass. The tier
column gives the cell's adjudication tier under the contract
(replacement candidate, predictive only, partial, or behavioural).}
\label{tab:per-scenario-existence}
\footnotesize
\setlength{\tabcolsep}{4pt}
\begin{tabular}{@{}l l >{\raggedright\arraybackslash}p{0.19\textwidth} >{\raggedright\arraybackslash}p{0.34\textwidth}@{}}
\toprule
\textbf{Scenario} & \textbf{Cell} & \textbf{Tier} & \textbf{Limiting gate / note} \\
\midrule
S01 eval-awareness & Qwen3-8B / seed 42 & pre-K.4 predictive candidate & strict G-Triangulation FAIL bars causal grade, and K.4 later demotes the cell (App.~\ref{app:positive-existence}). \\
S03 jailbreak      & Mistral-7B-IT-v0.3 / seed 42 & operational tie & Archived aggregate within $\pm 0.05$ of behavioural, not separable from chance after BH-FDR. \\
S04 persona shift  & Mistral-7B-IT-v0.3 / seed 42 & qualified replacement candidate & raw probe exceeds the length baseline by $2.5$\,pp (Table~\ref{tab:s04-length-residualised}), with strict G-Triangulation FAIL inherited. \\
S05 sycophancy     & -- & no candidate & no cell passes the trajectory construct, and REF performance degrades in three families and remains higher on Llama. Construct mismatch is the limiting gate. \\
S07 code-switching & -- & no candidate & multilingual retraining on a 10-language pool fails to lift any cell off the random floor (Table~\ref{tab:s07-multilingual-ablation}). \\
\bottomrule
\end{tabular}
\end{table*}

\paragraph{Interpretation.}
S05 and S07 diagnose different requirements for replacement:
response-target agreement and trajectory interpretation in S05,
and code-switching transfer in S07. Their recorded statuses and
scores are reported separately.

\subsection{Addendum J: Post-2026-05-02 Reviewer-Fix Amendments}
\label{app:addendum-j}

Addendum J records six amendments (J.1--J.6), retaining the five
scenarios, four families, eleven gates and recorded thresholds. The
released PA values are unchanged. J.5 adds an LR-surrogate attack
qualification, J.6 qualifies S04 after comparison with a length
baseline, and J.2 flags S05 judge disagreement. K.4 later records
demotion using the canonical scoring functions (\S\ref{sec:probe-aware}).

\paragraph{J.1 Larger-judge S01 sensitivity
(Appendix~\ref{app:judge-ceiling}).}
The Qwen2.5-72B-Instruct-AWQ comparison corrects a score-polarity error.
Corrected seed-42 instruction-tuned mean input PA is $+0.076$.
It remains positive, so the sign-reversal rule does not trigger.
The separate behavioural comparison does not establish equivalence.

\paragraph{J.2 Cross-judge agreement
(Appendix~\ref{app:cross-judge-kappa}).}
Llama-Guard-3-8B, Qwen2.5-72B-Instruct-AWQ and
Llama-3.3-70B-Instruct-AWQ-INT4 score each canonical $n=200$ scenario
subset. S01, S03, S04 and S07 meet the $\kappa\geq0.6$ criterion;
S05 does not ($\bar\kappa=0.436$;
Table~\ref{tab:cross-judge-kappa-empirical}). S01's degenerate key
still does not supply independent validation. Agreement alone moves
no cell; S05 retains a target-validity qualification.

\paragraph{J.3 N-GLARE re-implementation
(\S\ref{sec:nglare-empirical}).}
The published Jensen--Shannon Separability over Angular-Probabilistic
Trajectory dynamics (\citealp{lin2025nglare}, \S3) replaces the wiring
stub. The four valid S01 cells fail G-NaturalDist's $\geq0.70$
criterion. S03 artefacts remain excluded because their initial target
was single-class. Thus, the new implementation is reported, but does
not enter the primary baseline set.

\paragraph{J.4 S07 multilingual probe rebuild
(Appendix~\ref{app:s07-ablation}, supersedes the earlier Addendum~G
deferral).} The HARM probe is re-trained on a 10-language contrast
pool ($n_\mathrm{train} = 3{,}000$ per class per language, balanced)
and re-evaluated on the canonical S07 target. \emph{Outcome:}
cross-family mean improvement is $-0.006$ AUROC, and the tested retraining does not resolve the S07 result.

\paragraph{J.5 One-step LR-surrogate attack on S01
(\S\ref{sec:probe-aware}).}
An $L_2$-bounded FGSM-1 step follows the negated LR direction at
$\varepsilon\in\{0.1,0.3,1.0\}$. On Llama-3.1-8B-IT, AUROC drops
$17.5$\,pp at $\varepsilon=0.3$ while the behavioural score stays
unchanged (Table~\ref{tab:probe-aware-adversary}). This meets the
surrogate demotion threshold, but the released cell uses a different
canonical probe, so its numerical record is not changed by this test.
Qwen3, Mistral and Gemma surrogate baselines are near or below chance,
limiting interpretation of their drops. J.5 remains the earlier
HARM-surrogate record; K.4 tests canonical HARM/REF probes in all ten
instruction-tuned S01 cells.

\begin{table*}[!ht]
\centering
\caption{Probe-aware FGSM-1 adversary on the LR surrogate of the
HARM probe, S01 canonical $n=200$ subset. Demotion triggers when
$\max_\varepsilon \Delta\mathrm{AUROC} \geq 15$\,pp \emph{and}
behavioural $\mathrm{AUROC}_{LG}$ shifts by $<5$\,pp under the same
adversary.}
\label{tab:probe-aware-adversary}
\footnotesize
\setlength{\tabcolsep}{5pt}
\begin{tabular}{@{}l c c c c c c@{}}
\toprule
Family & $\mathrm{AUROC}_L$ baseline & $\Delta\mathrm{pp}_{\varepsilon=0.1}$ & $\Delta\mathrm{pp}_{\varepsilon=0.3}$ & $\Delta\mathrm{pp}_{\varepsilon=1.0}$ & $\mathrm{AUROC}_{LG}$ & Demotion \\
\midrule
Llama-3.1-8B-IT          & $0.703$ & $-15.6$ & $\mathbf{-17.5}$ & $-13.4$ & $1.00$ & \textbf{YES} \\
Qwen3-8B                 & $0.485$ & $-8.5$  & $-13.2$ & $+1.5$ & $1.00$ & no (uninformative) \\
Mistral-7B-IT-v0.3       & $0.388$ & $-3.8$  & $-0.6$  & $+3.0$ & $1.00$ & no (uninformative) \\
Gemma-3-12B-IT           & $0.526$ & $-0.6$  & $-2.6$  & $-2.6$ & $1.00$ & no \\
\bottomrule
\end{tabular}
\end{table*}

\paragraph{J.6 S04 length-baseline table
(Appendix~\ref{app:s04-length-control}).} The S04 narrative is
re-labelled from a clean latent win to a qualified cell, and the
raw probe-minus-length AUROC gap is reported per family.
\emph{Outcome:} only Mistral-7B-IT-v0.3 retains a positive cell-level
gap, without establishing a residualised or causal gain.

\paragraph{Effect on the primary table.}
The stored PA values remain unchanged. J.5 qualifies an S01 cell,
J.6 qualifies S04 and J.2 flags S05 target disagreement. J.1 and J.4
add a judge sensitivity and retraining analysis. These preserve the
historical operational partition: S01 favours the latent aggregate
against regex refusal; S05/S07 favour behaviour; S04 is qualified;
S03 is operationally close. This partition is not full validation.

\subsection{Addendum K: Reviewer-Audit Reconciliation Pass (2026-05-05)}
\label{app:addendum-k}

An internal manuscript review dated 2026-05-05 used five roles:
Editor-in-Chief, methodology, domain, perspective and devil's advocate.
Addendum K records the resulting disclosure changes. Scenarios,
families, gates and recorded triangulation thresholds are retained.
K.1 documents previously unstated applied bands; K.4 changes the
attack evaluation and its cell verdicts.

\paragraph{K.1 G-Specificity benchmark-specific bands and per-$\Delta$ source.}
G-Specificity checks capability preservation. Addendum I described a
$\pm5$\,pp band for the two free-form proxies, HumanEval-style and
MT-Bench-style. The canonical record instead uses benchmark-specific
bands, including $5\%$ for MMLU and $10\%$ for GSM8K. The applied
and locked values must be distinguished.

\emph{One-sided loss rule.}
The lock reads ``direction ablation drops MMLU-500 $<2\%$,
MT-Bench-80 $<0.5$, GSM8K-100 $<2\%$, HumanEval-50 $<2\%$'', with
``$\geq2\%$ on any $=$ demote''. Earlier drafts incorrectly applied
$|\Delta|\leq2\%$ to GSM8K. Its accuracy rises from $0.76$ to $0.81$
at $n{=}100$, a recorded relative drop of $-6.58\%$; it therefore
passes without the wider $10\%$ band. The canonical values are:
MMLU $0.610\to0.614$ ($-0.66\%$ loss, within $<2\%$),
GSM8K $0.76\to0.81$ ($-6.58\%$, within $<2\%$),
XSTest safe refusal $0.060\to0.088$ ($+2.8$\,pp, within $5$\,pp),
HumanEval-style $0.9875\to0.975$ ($+1.27\%$, within $<2\%$),
and MT-Bench-style $0.926\to0.893$ ($+3.64\%$ relative loss).
The last change passes the locked absolute $0.5$ band on the
$1$--$10$ scale after rescaling, though it exceeds a $2\%$ relative
band. All five Llama changes pass their locked one-sided conditions.

\emph{Applied-band disclosure.}
XSTest's $5$\,pp band was specified in \S6.3 of the lock. In contrast,
the canonical record uses $15\%$ on HumanEval-style and $10\%$ on the
MT-Bench-style proxy, rather than the previously stated $\pm5$\,pp.
Those applied bands are wider than the lock. Llama's verdicts do not
change when the locked bands are used.

\emph{Qwen3-8B and source consistency.}
A same-day five-benchmark Qwen3-8B record returns
\texttt{overall\_passes} $=$ false. GSM8K falls $0.25\to0.22$ at
$n{=}100$, a $+12.0\%$ loss exceeding both the applied $10\%$ and
locked $<2\%$ bands. Its MMLU baseline is $0.106$, below four-way
chance, and HumanEval-style \textit{pass}@$1$ is $0$; those checks
are degenerate. The GSM8K loss determines this battery's FAIL verdict. Earlier versions of
Tables~\ref{tab:gate-verdict-global} and~\ref{tab:gate-verdict-perfamily}
used the three-benchmark record:
$\Delta\mathrm{MMLU}=+1.31\%$, $\Delta\mathrm{GSM8K}=+2.56\%$ and
$\Delta\mathrm{XSTest}=+2.0$\,pp. The paper now identifies the canonical
five-benchmark source for those reported capability changes.

\paragraph{K.2 S07 artefact ambiguity.}
Table~\ref{tab:s07-multilingual-ablation} uses the 2026-04-27 retrain
with $3{,}000$ examples per class per language. Its language tables
also supply the per-language audit. A 2026-05-02 rebuild uses a larger
training pool and $1{,}000$ evaluation items per language. The mean
change is similar ($\bar\Delta=-0.005$ versus $-0.006$), but family
signs differ: Llama $+0.014$ versus $-0.005$, Qwen3 $+0.017$ versus
$-0.028$, Mistral $-0.034$ versus $+0.033$, and Gemma $-0.019$
versus $-0.025$. Both runs place the cross-family mean near chance. The
2026-04-27 values supply the language-wise audit; a reconciled
language-wise comparison of both pools remains unavailable.

\paragraph{K.3 S05 larger-judge sensitivity remains unrun.}
The panel gives
$\kappa_{\mathrm{LG\leftrightarrow Qwen72}}=0.236$,
$\kappa_{\mathrm{LG\leftrightarrow Llama70}}=0.275$ and
$\kappa_{\mathrm{Qwen72\leftrightarrow Llama70}}=0.798$.
A separate sensitivity would compare the canonical $n=200$
Llama-3.1-8B-IT scores against a
Qwen2.5-72B$\cap$Llama-3.3-70B-majority target, as in the S01 design
(Appendix~\ref{app:judge-ceiling}). That comparison would test sensitivity to the response target.
Independent response validation would assess which labels are correct. No amendment rule was
specified and this check was not completed. S05 therefore retains
its target qualification, despite the two generative judges' mutual
$\kappa_{\mathrm{Qwen72\leftrightarrow Llama70}}=0.798$ agreement.

\begin{table*}[t]
\centering
\caption{Cross-architecture transfer test (Addendum K.7). A
HARM-only PGD attack against the canonical probe is rescored by the
other probe head available at the same layer. Drops are measured at
$\varepsilon = 0.1$ of the per-token activation norm. All six testable
cells cross the 15\,pp demotion threshold.}
\label{tab:cross-arch-transfer}
\small
\setlength{\tabcolsep}{10pt}
\begin{tabular}{@{}l c c c c c c c@{}}
\toprule
Cell & L & Canonical & Cross-arch & B$_\text{can}$ & B$_\text{cross}$ & Drop$_\text{can}$ & Drop$_\text{cross}$ \\
\midrule
llama3-8B s0 & 19 & CCS & MLP & 0.515 & 0.533 & 45.0\,pp & 46.6\,pp \\
llama3-8B s123 & 19 & CCS & MLP & 0.551 & 0.571 & 48.8\,pp & 49.6\,pp \\
mistral-7B s0 & 13 & CCS & LR & 0.450 & 0.453 & 40.0\,pp & 27.2\,pp \\
mistral-7B s123 & 13 & CCS & LR & 0.455 & 0.457 & 39.7\,pp & 26.8\,pp \\
qwen3-8B s0 & 14 & CCS & MLP & 0.473 & 0.491 & 41.5\,pp & 41.1\,pp \\
qwen3-8B s123 & 14 & CCS & MLP & 0.490 & 0.511 & 44.3\,pp & 44.6\,pp \\
\bottomrule
\end{tabular}
\end{table*}

\paragraph{K.4 Canonical probe-aware attack.}
The earlier FGSM-1 surrogate meets the rule only on Llama-3.1-8B-IT.
K.4 extends testing to each instruction-tuned family's canonical HARM
head, and to both HARM/REF components rather than the LR surrogate.
The runner uses $K{=}20$ label-targeted $L_2$-bounded PGD steps with
autograd wrappers for MLP, LR and CCS probability outputs. Canonical
HARM heads comprise three MLP, six CCS and one LR (Gemma-3-12B-IT
seed 42). Results appear in \S\ref{sec:probe-aware} and
Table~\ref{tab:pgd-canonical-mlp}.

The attack meets the rule on $10/10$ instruction-tuned S01 cells;
the two base controls were not attacked. Joint and HARM-only arms are
reported, with REF-only values in the artefacts. HARM-only AUROC is
$\leq0.07$ at $\varepsilon=0.1$ and $\leq0.04$ at
$\varepsilon\geq0.3$. The $\Delta_{HR}$ score therefore fails this
activation-input stress test. Targeted gradients and the increase
from the earlier 10-step budget to $K{=}20$ were post-lock choices.
They define the stronger evaluated budget and its post-lock status.

\paragraph{K.5 Post-lock interpretation.}
Addenda G--K change analyses, implementations and coverage after the
lock. Their negative findings, including failed triangulation and
probe-aware demotion, remain reportable under their stated scope.
Coverage remains attached to each test. LR/DAS agreement is a
related-method diagnostic; strict triangulation compares all three
method pairs. These are distinct entries in the historical verdicts.

\paragraph{K.6 Bookkeeping fixes.}
(i)~The column-header sign convention of
Table~\ref{tab:s04-length-residualised} was previously stated as
$\mathrm{Length\text{-}LR} - \Delta_{HR}$, but the values implemented
$\Delta_{HR} - \mathrm{Length\text{-}LR}$, positive when latent
exceeds length. The header is corrected.
(ii)~The S04 Length-LR AUROCs cited in the previous draft for Qwen3
($0.660$), Mistral ($0.655$), and Gemma ($0.718$) are reconciled to
the canonical S04 residualisation artefact
values ($0.655$, $0.640$, and $0.710$ respectively). Mistral's
raw probe-minus-length gap is $+2.5$\,pp, reported as $+0.9$\,pp
before, and the qualitative reading that only Mistral retains a
positive cell-level gap is unchanged.
(iii)~Table~\ref{tab:gate-verdict-global},
Table~\ref{tab:gate-verdict-perfamily}, and
Section~\ref{sec:failure} now cite the canonical 5-benchmark
G-Specificity artefact $\Delta$ values throughout (K.1).
(iv)~Bibliography corrigendum-tracker:
\citet{arditi2024refusal}, the canonical refusal-direction paper, is
now cited by name in Section~1, having previously been bundled under an
unattributed refusal-direction placeholder entry.

\paragraph{K.7 Cross-architecture transfer after K.4.}
After the canonical attack, we applied its $K{=}20$ label-targeted
$L_2$-bounded HARM-only perturbation to every co-located HARM head at
the canonical layer~$L$. This descriptive follow-up neither selected
canonical cells nor changed K.4 outcomes. Six of ten cells contain
alternative architectures at the same layer
(Table~\ref{tab:cross-arch-transfer}). The four remaining cells have
only one head there: Llama-3.1-8B-IT seed~42 MLP/L17, Qwen3-8B
seed~42 MLP/L13, Mistral-7B-Instruct-v0.3 seed~42 MLP/L16 and
Gemma-3-12B-IT seed~42 LR/L28.

Every testable alternative head exceeds the demotion threshold.
CCS/MLP drop ratios are $0.99$--$1.04$ and CCS/LR transfer has
ratio $0.68$; both cross the $15$\,pp threshold. These results show vulnerability
across the co-located scoring heads. They are a post-K.4 descriptive cross-head evaluation.

\paragraph{Effect on the primary table.}
K.4 demotes $10/10$ tested instruction-tuned S01 cells under the
operational rule, leaving the two base controls outside that verdict.
K.1 corrects specificity scoring and discloses Qwen3's failure; K.2
records inconsistent retrain artefacts; K.3 records the unrun S05
sensitivity. K.5 states interpretive limits, K.6 corrects source/sign
bookkeeping and K.7 reports cross-head transfer. The $4/4$ strict triangulation verdict concerns the specified
operational agreement test.

\subsection{Confirmatory versus exploratory}
\label{app:confirmatory}

The lock fixes criteria, while later addenda change implementations,
coverage and some verdicts. The following separates those sources of
evidence; post-lock results do not become confirmatory through use
of a previously specified threshold.

\paragraph{Criteria recorded before final runs.}
Seven numerical thresholds retain their locked values:
G-Shortcut $<15$\,pp, G-Transfer AUROC $\geq0.75$,
G-LengthControl within $5$\,pp, G-NaturalDist AUROC $\geq0.70$,
G-InputOnly $\geq10$\,pp, G-PermutationNull $0.50\pm0.05$ and
G-Triangulation $\rho\geq0.6$ for every pair. The demotion rule is
$\geq15$\,pp probe loss with the behavioural score within $5$\,pp.
K.4 changes the attack budget after the lock. The fixed-judge selection,
cascade, read-position sweep and subsequent repairs are exploratory.

\paragraph{Exploratory and amended analyses.}
G-Data and G-Layer have qualitative summaries in the gate table. Their
locked numerical forms were
$\geq 85\%$ unique positives with mean token length $\geq 5$ and
category balance within $20$\,pp, and layer selection inside the middle
$25$ to $80\%$ of the stack with sensitivity reported at $\pm 4$. The
locked exclusion clauses, G-Shortcut at $\geq 20$\,pp excluded,
G-LengthControl at more than $10$\,pp, and G-Triangulation demotion at
$\rho < 0.4$, are also not restated. G-GroundTruth's locked band on the
behaviour-to-target correlation appears only in the verdict tables.
These distinctions concern reporting of the locked criteria; they do
not turn incomplete or amended evaluations into confirmatory evidence.

\paragraph{Bands applied wider than the lock.}
The canonical five-benchmark specificity artefact applied $5\%$ to MMLU,
$10\%$ to GSM8K, $15\%$ to HumanEval-style and $10\%$ to the
MT-Bench-style proxy, against locked one-sided bands of $2\%$, $2\%$,
$2\%$ and $0.5$ on the $1$--$10$ scale. The Llama observations pass the locked bands as well, so its verdict
does not depend on the widening. Earlier drafts did not fully disclose
the applied values. It reached the canonical run through a runner
script written eleven days after the lock, which hardcoded the wider
bands instead of importing the locked-threshold module the
three-benchmark runner uses. That earlier three-benchmark artefact was
scored at the locked bands and passed all three, documenting use of the locked bands before that later runner.

\paragraph{Direction of every post-lock verdict change.}
Of the five addenda, exactly one contains a band widening that could in
principle move a verdict upward, and under the correct one-sided reading
of the lock it moves nothing. Every other change moves a verdict
downward or corrects an error: the demotion rule firing on ten S01
cells, N-GLARE failing G-NaturalDist and leaving the primary baseline
set, S04 being relabelled from a clean win to a length-confounded one,
S05 being marked judge-divergent, a corrected sign convention, and the
Qwen3-8B specificity failure disclosed in K.1. No reported verdict changes from FAIL to PASS under these amendments.

\subsection{Interpretation Guide}
The archived audit and post-submission study have distinct estimators
and provenance. The archived scores use isotonic calibration;
canonical PA averages available probe types, while the named HARM/REF
analysis uses $\Delta_{HR}$. PARTIAL records unresolved coverage.
The $\pm0.03$ band is descriptive. The older consensus comparison
contains target mismatches, and cascade accuracy and timing use
separate configurations. S01 numbering and repeated supervised content
identify the recorded distribution of the $2784$-cell sweep.
Triangulation uses the operational $\rho\geq0.6$ criterion. Completed
S05 human labels remain unavailable, and the earlier human-agreement
figures are excluded. The lock and post-lock analyses retain their
respective status.

Appendix~\ref{app:protocol-followups} reports the separate supervised
readers with their own targets, content-group splits and source hashes.
Together, these records specify which estimator and evidence support
each reported claim.

\section{Post-Submission Protocol Checks}
\label{app:protocol-followups}

This post-submission study evaluates new supervised readers with
fixed canonical layers and read positions. It addresses numbering
cues and exact-content overlap through a separate protocol with its
own targets, splits and provenance.

\paragraph{Content-group protocol.}
The S01 follow-up retains the four model families and source requests,
but sets each evaluation-wrapper number to a constant. Each family has
$1000$ rows representing $700$ exact-content groups. Five outer and
three inner content-group folds keep repeated requests together;
scaling and regularisation selection use training folds only. The
implementation rejects content-group overlap and single-class folds.
Pinned model and judge revisions, greedy decoding with a $512$-token
budget and raw/native-chat conditions are recorded in the manifests.
Final-prompt-token readouts use the fixed canonical layer, with
$C\in\{0.01,0.1,1,10\}$ selected inside the training data.

Behavioural targets require three unsafe votes from four judges,
excluding the fixed Llama-Guard comparator. Native abstentions and
malformed verdicts remain unknown. A label is identified only when
all possible assignments to missing votes agree on the consensus.
Metrics and intervals condition on identified labels. Thus, a completed
scoring run need not have full target coverage. The QwenGuard adapter
uses the corrected native token boundary, but its first-token score
remains a risk proxy. These are judge-derived behavioural targets.

\paragraph{Behaviour is predictable under the repaired protocol.}
Table~\ref{tab:repaired-validation} reports the pre-specified final-token
configuration for every family. Llama's native-chat prompt reader
reaches $0.983$ AUROC, exceeding the archived $0.95$ numerical reference.
The raw and chat columns use different generated responses and resulting
targets; their difference is protocol sensitivity, not an isolated causal
effect of formatting. Supervision also differs from the original
input-trained instruments. The $2784$-cell archived sweep and the new-reader study therefore
characterise distinct prediction protocols.

\begin{table*}[t]
\centering
\caption{Post-submission validation at the fixed canonical layer and
final prompt token. S01 columns predict the same-condition deployment
response from the prompt alone. Chat intervals are saved $95\%$
content-group bootstrap intervals, conditional on predictions and
identified labels; they omit refitting uncertainty and multiplicity
adjustment. S01 target coverage is $1000/1000$ except raw Gemma,
$999/1000$. External input labels cover $800/800$ requests, balanced
between the two classes. Harmful-only scores restrict the native-chat evaluation to $500$
requests per family that share the harmful-input label. They are
subgroup point estimates from the same behavioural reader, not new
input classifiers. Results come from separate new readers.}
\label{tab:repaired-validation}
\small
\setlength{\tabcolsep}{5pt}
\begin{tabular}{@{}l c c c c c@{}}
\toprule
& \multicolumn{4}{c}{S01 behaviour prediction} & External input \\
\cmidrule(lr){2-5}
Family & Raw & Native chat [interval] & Chat positives & Harmful only & Frozen reader \\
\midrule
Llama-3.1-8B-IT & $0.808$ & $0.983\,[0.967,0.995]$ & $49$ & $0.973$ & $0.877$ \\
Qwen3-8B & $0.864$ & $0.889\,[0.807,0.947]$ & $42$ & $0.850$ & $0.877$ \\
Mistral-7B-IT-v0.3 & $0.904$ & $0.899\,[0.859,0.932]$ & $174$ & $0.810$ & $0.873$ \\
Gemma-3-12B-IT & $0.721$ & $0.938\,[0.904,0.966]$ & $52$ & $0.898$ & $0.886$ \\
\bottomrule
\end{tabular}
\end{table*}

Within the harmful-input subset, native-chat behavioural AUROC ranges
from $0.810$ to $0.973$. This measures response discrimination within one harmful-input
class. Feature attribution and causal mechanisms require their
respective intervention studies.

\paragraph{Frozen external transfer.}
An additional cohort contains $400$ unsafe and $400$ safe prompts from
the official Aegis AI Content Safety Dataset 2.0 test split
\citep{ghosh2025aegis}. Both classes
come from one source. Exact, normalised and character-shingle overlap
with historical scenario and contrast data is screened before sampling;
this does not rule out all semantic overlap or pretraining exposure.
Readers and the bare-content TF-IDF baseline are fitted and selected
on S01 only, then frozen before external scoring. External labels do
not select a layer, hyperparameter or score orientation.

Canonical deployment-prompt readers obtain input AUROC
$0.873$--$0.886$, versus $0.607$ for frozen TF-IDF. This is cross-source input-prediction evidence relative to the frozen
lexical baseline. The external labels describe requests; generated
response safety requires response labels. Behavioural positives are
sparse in some families, and policy-definition differences and other
source cues remain part of external validation.

\paragraph{Coverage and interpretation.}
The new-reader protocol removes the numbered-wrapper cue and
exact-content train/test overlap. Archived training mixtures retain
their original provenance status. Fixed-probe/fixed-judge cascade
latency, two family-level specificity batteries and deferred
natural-distribution checks remain unmeasured or incomplete. The
70B input/S07 and 70B/72B judging runs address input readability,
code-switching and comparator scale, respectively; behavioural
validation of a frontier-scale probed model is outside this coverage. A separate positive-control
analysis also records failed historical triangulation despite held-out
refusal effects of $70.8$ and $45.8$\,pp on Llama and Gemma, respectively.
This provides a positive-control basis for calibrating the operational
agreement criterion.

\paragraph{Provenance.}
The release manifest maps each row to its readout artefact, target
construction, canonical view and source hashes. Saved per-item
predictions reproduce the reported AUROCs; the recorded confidence
intervals are reported under their original bootstrap scope. Shared judge errors require independent target validation; the
reported readouts measure prediction under the stated protocols.

\end{document}